\documentclass[10pt,journal,compsoc]{IEEEtran}

\ifCLASSOPTIONcompsoc
  \usepackage[nocompress]{cite}
\else
  \usepackage{cite}
\fi

\ifCLASSINFOpdf
\else
\fi

\usepackage{algorithm}
\usepackage{algorithmic}
\usepackage{booktabs}       
\usepackage{subcaption}
\usepackage{amsmath}
\usepackage{amsthm}

\usepackage{amssymb}
\usepackage{newfloat}
\usepackage{listings}
\usepackage{bbm}
\usepackage{multirow}
\usepackage{array}
\usepackage{enumitem}
\usepackage{graphicx}
\usepackage{comment}
\usepackage{url}

\begin{document}
\begin{sloppy}
\title{Revisiting Transformation Invariant Geometric Deep Learning: An Initial Representation Perspective}

\author{Ziwei~Zhang,~\IEEEmembership{Member,~IEEE,}
		Xin~Wang,~\IEEEmembership{Member,~IEEE,}
		Zeyang~Zhang,
        Peng~Cui,~\IEEEmembership{Senior Member,~IEEE,}
        and~Wenwu~Zhu,~\IEEEmembership{Fellow,~IEEE}%
\IEEEcompsocitemizethanks{\IEEEcompsocthanksitem All authors are with the Department of Computer Science and Technology at Tsinghua University, Beijing 100084, China.  \protect\\
Email: \{zwzhang,xin\_wang,cuip,wwzhu\}@tsinghua.edu.cn, \protect\\zy-zhang20@mails.tsinghua.edu.cn}
\thanks{Manuscript received April 19, 2005; revised August 26, 2015.}}

\markboth{Journal of \LaTeX\ Class Files,~Vol.~XX, No.~X, XXX~2022}%
{Shell \MakeLowercase{\textit{et al.}}: Bare Demo of IEEEtran.cls for Computer Society Journals}

\newtheorem{remark}{Remark}
\newtheorem{theorem}{Theorem}
\newtheorem{definition}{Definition}

\newcolumntype{P}[1]{>{\centering\arraybackslash}p{#1}}

\newcommand{\tabincell}[2]{\begin{tabular}{@{}#1@{}}#2\end{tabular}}

\newcommand{\model}{{TinvNet }}
\newcommand{\modelns}{{TinvNet}}

\IEEEtitleabstractindextext{%
\begin{abstract}
Deep neural networks have achieved great success in the last decade. When designing neural networks to handle the ubiquitous geometric data such as point clouds and graphs, it is critical that the model can maintain invariance towards various transformations such as translation, rotation, and scaling. Most existing graph neural network (GNN) approaches can only maintain permutation-invariance, failing to guarantee invariance with respect to other transformations. Besides GNNs, other works design sophisticated transformation-invariant layers, which are computationally expensive and difficult to be extended. In this paper, we revisit why general neural networks cannot maintain transformation invariance. Our findings show that transformation-invariant and distance-preserving initial point representations are sufficient to achieve transformation invariance rather than needing sophisticated neural layer designs. Motivated by these findings, we propose Transformation Invariant Neural Networks (\modelns), a straightforward and general plug-in for geometric data. Specifically, we realize transformation invariant and distance-preserving initial point representations by modifying multi-dimensional scaling and feed the representations into existing neural networks. We prove that \model can strictly guarantee transformation invariance, being general and flexible enough to be combined with the existing neural networks. Extensive experimental results on point cloud analysis and combinatorial optimization demonstrate the effectiveness and general applicability of our method. We also extend our method into equivariance cases. Based on the results, we advocate that \model should be considered as an essential baseline for further studies of transformation-invariant geometric deep learning.
\end{abstract}

\begin{IEEEkeywords}
Transformation Invariance, Geometric Deep Learning, Combinatorial Optimization, Point Cloud, Graph Neural Network
\end{IEEEkeywords}}

\maketitle

\IEEEdisplaynontitleabstractindextext
\IEEEpeerreviewmaketitle
\IEEEraisesectionheading{\section{Introduction}\label{sec:introduction}}

\IEEEPARstart{D}{eep} neural networks~\cite{lecun2015deep} have achieved enormous successes in many fields such as computer vision~\cite{krizhevsky2012imagenet}, natural language processing~\cite{devlin2019bert}, and game playing~\cite{mnih2015human}. On the other hand, geometric data, such as graphs or point clouds, is ubiquitous in practice, ranging from molecules in proteins to 3D objects. Compared with grid-structured data such as acoustics, images, or videos, geometric data poses more challenges for designing suitable neural networks due to the irregular structure~\cite{bronstein2017geometric}.

To handle geometric data effectively, one critical \emph{inductive bias} is to design invariant and equivariant models with respect to various transformations such as permutation, translation, rotation, reflection, and scaling. Take 3D object recognition in point cloud analysis as an example. The shape of an object is invariant to isometric transformations such as translation, rotation, and reflection. For the classic NP-hard travelling salesman problem (TSP), the solution of a TSP is invariant to isometric and the scaling transformations of the coordinates. Maintaining invariance and equivariance with respect to these transformations can enhance the generalization ability, robustness, and interpretability of neural networks in handling geometric data~\cite{bronstein2017geometric,cohen2016group}. However, designing transformation invariant neural networks for geometric data poses great challenges. For instance, though convolutional neural networks (CNNs) are known to enjoy and benefit from translation equivariance in handling images and videos~\cite{ravanbakhsh2017equivariance}, it is non-trivial to extend such merit to geometric data since there is no grid structure.

As an emerging type of neural networks to process geometric data, graph neural networks (GNNs) have been shown effective in a wide range of geometric applications such as protein interaction prediction~\cite{zitnik2017predicting}, point cloud analysis~\cite{wang2019dynamic}, combinatorial optimization~\cite{bengio2020machine}, etc. Maintaining permutation-equivariance is a crucial reason behind the success of GNNs, i.e., if we randomly permute the IDs of nodes, the representations produced by GNNs are permuted accordingly. By adopting the message-passing framework~\cite{gilmer2017neural}, most GNNs can easily satisfy permutation-equivariance~\cite{maron2018invariant,keriven2019universal}. However, most GNNs largely ignore other transformations mentioned above, e.g., rotation and scaling. Ideally, GNNs should be able to produce equivariant or invariant representations when geometric data is transformed. But the existing GNNs consider transformed data as independent samples, failing to produce desired representations. 

Other attempts to alleviate the problems caused by transformations include data augmentation and manually extracting transformation invariant features such as the distance and angle between geometric objects~\cite{klicpera2019directional,chen2019clusternet}. Due to the massive number of possible transformations, data augmentation cannot guarantee effectiveness and works poorly in practice. Meanwhile, manually designed features can only preserve a limited amount of information. Some works have designed specific neural network layers targeting certain transformations such as rotations~\cite{thomas2018tensor,fuchs2020se}. Though having made some progresses, these methods usually resort to complicated group theories and geometric analysis, suffering from being computationally expensive and difficult to be extended~\cite{fuchs2020se}. For example, when incorporating the attention mechanism~\cite{vaswani2017attention} into an existing rotation-equivariant network named tensor field network~\cite{thomas2018tensor}, great efforts are needed to redesign all operators~\cite{fuchs2020se}.

To address this problem, we first revisit why the existing neural networks cannot maintain transformation invariance when handling geometric data. We find that transformation-invariant and distance-preserving initial point representations are sufficient to achieve transformation invariance rather than needing to design sophisticated neural network layers as proposed in the existing methods. Motivated by these findings, we propose Transformation Invariant Neural Network (\modelns), a straightforward and general plug-in for geometric data. Specifically, we realize transformation invariant and distance-preserving initial point representations by modifying multi-dimensional scaling (MDS), a classical dimensionality reduction technique. We then feed the initial representation into neural networks. We prove that such a simple mechanism can strictly guarantee transformation invariance. Besides, since \model is a general framework compatible with existing neural networks, it is flexible to be combined with various architectures such as different GNN variants or other neural networks for geometric data. We further provide extending our method to equivariance cases in Appendix~\ref{sec:equ}.

We conduct extensive experiments on tasks including point cloud analysis and combinatorial optimization. The results show that \model is indeed strictly invariant to various transformations such as translation, rotation, reflection, and scaling. In the rotation transformation setting of point cloud analysis, \model combined with DGCNN~\cite{wang2019dynamic}, a well-known non-rotation-invariant GNN model, outperforms or matches the performance of various recently proposed models specifically designed to be rotation-invariant. Our proposed model also significantly outperforms a recent learning-based model for combinatorial problems when transformations are involved. Besides, thanks to the simple and general mechanism of \modelns, it is easily compatible with multiple architectures. Based on the experimental results, we advocate that \model should be considered a new starting point and an essential baseline for further studies of transformation-invariance on geometric data. Our contributions are summarized as follows:
\begin{itemize}[leftmargin = 0.4cm]
	\item We revisit transformation invariant geometric deep learning and show that transformation-invariant and distance-preserving initial point representation is sufficient to solve the problem. 
	\item Motivated by the findings, we propose Transformation Invariant Neural Network, a straightforward and general plug-in that is proved to be strictly transformation-invariant as well as general and flexible to combine with various neural networks.
	\item Extensive experimental results on tasks including point cloud analysis and combinatorial optimization demonstrate the efficacy and general applicability of our model.
\end{itemize}
The rest of the paper is organized as follows. In Section 2, we review related work. The problem formulation is introduced in Section 3. We revisit the transformation invariance problem and propose the \model model in Section 4. We report experimental results in Section 5, and conclude our paper in Section 6.

\section{Related Work}
In this section, we first review GNNs and their permutation equivariance and invariance properties. Then, we review other invariance for geometric deep learning.
\subsection{Graph Neural Networks and Permutation Equivariance/Invariance}
GNNs are one emerging type of neural networks to process geometric data. Early GNNs such as recursive architectures~\cite{gori2005new,scarselli2008graph} and contextual realizations~\cite{micheli2009neural} predate the rise of deep neural networks. Nevertheless, it is not until the deep learning era that GNNs  gain popularity. Recent advances in GNNs can be broadly categorized into spectral approaches~\cite{bruna2014spectral,xu2018graph,defferrard2016convolutional} and spatial approaches~\cite{niepert2016learning,duvenaud2015convolutional,li2016gated}. For spectral approaches, graph signal processing techniques~\cite{shuman2013emerging,ortega2018graph} are adopted to process graph data in the spectral domain. For spatial approaches, the neural networks directly work on the connectivity patterns of graphs. Due to the efficiency and effectiveness, the message-passing framework~\cite{gilmer2017neural} is a de facto standard in recent GNN designs, including GCN~\cite{kipf2017semi}, GraphSAGE~\cite{hamilton2017inductive}, GAT~\cite{velivckovic2018graph}, JK-Nets~\cite{xu2018representation}, GIN~\cite{xu2019powerful}, and Graph Nets~\cite{battaglia2018relational} as particular instantiations.

One fundamental property of the message-passing GNNs is permutation-equivariance, i.e., the node representations are not dependent on node IDs. For example, many studies~\cite{li2018deeper,morris2019weisfeiler,maron2019provably} analyze the connection between GNNs and the Weisfeiler-Lehman (WL) algorithm~\cite{shervashidze2011weisfeiler} of graph isomorphism tests. Since the WL algorithm is strictly permutation-equivariant, GNNs also need to be strictly permutation-equivariant to mimic WL algorithms. By applying permutation-invariant pooling layers~\cite{zhang2018end,murphy2019relational,ying2018hierarchical} on permutation-equivariant node representations, permutation-invariant graph representations can be obtained~\cite{maron2018invariant,keriven2019universal}. Permutation is orthogonal to the similarity transformations studied in this paper. By adopting permutation-equivariant GNNs as backbones, our model can also satisfy permutation equivariance and invariance.

\subsection{Other Invariance for Geometric Deep Learning}
Rotation-invariant and equivariant neural networks for geometric data have been studied previously~\cite{cohen2018spherical,weiler20183d,cohen2016group}, particularly in point clouds~\cite{chen2019clusternet,thomas2018tensor,zhang2019rotation,kim2020rotation}. Most of these methods do not consider other transformations such as scaling, translation, or reflection. Besides, these methods design sophisticated neural network layers inspired by group theories and geometric analysis to guarantee rotation-invariance and equivariance. In general, these methods are complicated, computationally expensive, and difficult to be extended. In comparison, our proposed method is simple and straightforward. We also empirically compare our method with these methods in Section~\ref{sec:pointcloud}.

A scale-invariant GNN is proposed in~\cite{tang2020towards} to handle different scales of node features. However, it cannot handle other transformations such as translation and rotation. Very recently, geometrically invariant and equivariant GNNs began to receive attentions~\cite{han2022geometrically}. For example, IsoGCN~\cite{horie2021isometric} is proposed to handle isometric transformations and EGNN~\cite{satorras2021n} is proposed to handle various transformations by designing sophisticated message-passing functions. In general, these methods also design sophisticated neural network layers, e.g., certain types of message-passing, to realize invariance and equivariance. In comparison, our method studies the problem from another perspective and is more straightforward and compatible with existing neural networks.

There are also recent works combining eigen-analysis and neural networks for geometric learning called intrinsic coordinations. For example, IEConv~\cite{hermosilla2020intrinsic} adopts one extrinsic and two intrinsic distances and designs a new convolution operator for protein modeling. Koestler et al.~\cite{koestler2022intrinsic} proposes an intrinsic neural field method for shapes. \model can also be regarded as a type of intrinsic coordinates. In comparison, our method is more straight-forward and compatible (only operating in the initial representation), empirically more effective, and fully distance-preserving. %

\section{Problem Formulation}
In this section, we introduce notations and preliminaries of transformation invariance and similarity transformation. We summarize notations in Table~\ref{tab:notation}.

\begin{table}[t]
	\caption{A Summary of Notations}
	\label{tab:notation}
	\begin{footnotesize}
		\centering
		\begin{tabular}{l | l } \toprule
			Symbol              & Meaning               \\ \midrule
			$\mathcal{V}=\left\{v_1,...,v_N\right\}$ & The set of points     \\
			$\mathbf{F}\in\mathbb{R}^{N \times d}$        & The coordinate matrix \\
			$\mathcal{D}(\cdot,\cdot), \mathbf{D}$ & The distance metric and distance matrix \\
			$\mathcal{G}=\left(\mathcal{V},\mathcal{E}\right)$ & A graph \\
			$\mathbf{W}$ & Learnable parameters in the neural network \\
			$\mathcal{T}(\cdot)$ & A similarity transformation \\
			$c,c^\prime$                & Scaling constants \\ 
			$\mathbf{H}^{(l)}$ & The point representation in the $l^{th}$ layer \\
			$\mathbf{H}^{(0)} = \mathcal{P}(\mathbf{F})$ & The initial representation and mapping function \\  
			$\mathbf{H} = \mathbf{H}^{(L)}$ & The final point representation \\
			$\mathbf{\Lambda},\mathbf{X}$ & Eigenvalues and the corresponding eigenvectors \\
			$\mathbf{S} $ & The similarity matrix \\
			$\mathbf{I}_N,\mathbf{1}_N$ & $N\times N$ identity matrix/matrix of ones 
			\\ \bottomrule
		\end{tabular}
	\end{footnotesize}
\end{table}

We consider each geometric data instance as a collection of points $\mathcal{V} = \left\{v_1, v_2,...,v_N\right\}$ with $N$ denoting the number of points. The points have a coordinate matrix $\mathbf{F} \in \mathbb{R}^{N \times d}$ with $d$ denoting the dimensionality. Denote by $\mathbf{F}_{i,:}$, $\mathbf{F}_{:,j}$, and $\mathbf{F}_{i,j}$, the $i^{\text{th}}$ row, $j^{\text{th}}$ column, and an element of the matrix, respectively. $\mathbf{F}_{i,:}$ is the coordinate of point $v_i$. We denote a symmetric distance metric associated with the coordinates as $\mathcal{D}(\cdot,\cdot)$. In this paper, we assume the metric is the Euclidean distance by default. There is a graph $\mathcal{G} = \left( \mathcal{V}, \mathcal{E} \right)$ to describe the relationships between points, where $\mathcal{E} \subseteq \mathcal{V} \times \mathcal{V}$ is a set of edges. The graph can be provided in the data or constructed from the coordinates, e.g., the k-nearest neighbors graph. We denote the adjacency matrix of the graph as $\mathbf{A}$. $\mathcal{N}(i) = \left\{v_j: \left(v_i,v_j\right) \in \mathcal{E}\right\}$ is the neighborhood of $v_i$. 

Neural networks for geometric data usually aim to learn representation $\mathbf{H}$ for the points using the coordinate matrix $\mathbf{F}$ and the graph $\mathbf{A}$. We generally denote such neural networks as\footnote{For non-graph-based neural networks for geometric data, $\mathbf{A}$ can be removed from Eq.~\eqref{eq:geomnn}. For notation convenience, we keep $\mathbf{A}$ in Eq.~\eqref{eq:geomnn}.}
\begin{equation}\label{eq:geomnn}
	\mathbf{H} = \rm{NN}\left(\mathbf{F},\mathbf{A}; \mathbf{W}\right),
\end{equation}
where $\mathbf{W}$ are learnable parameters. We mainly study how to maintain invariance with respect to different transformations applied to the geometric data.
\begin{definition}[Transformation Invariance]\label{def:inv}
	For a given transformation $\mathcal{T}(\cdot):\mathbb{R}^d \rightarrow \mathbb{R}^{d}$, a neural network following Eq.~\eqref{eq:geomnn} is transformation invariant if $\forall$ $\mathbf{F}$, $\mathbf{A}$, $\mathbf{W}$, the following equation holds:
	\begin{equation}
		\rm{NN}\left(\mathcal{T}(\mathbf{F}),\mathbf{A};\mathbf{W}\right) = \rm{NN}\left(\mathbf{F},\mathbf{A};\mathbf{W}\right),
	\end{equation}
	i.e., the model outputs identical point representations after the transformation.
\end{definition}
Transformation invariance in Definition~\ref{def:inv} is compositional, i.e., if a model is invariant with respect to both $\mathcal{T}_1(\cdot)$ and $\mathcal{T}_2(\cdot)$, the model is also invariant with respect to $\mathcal{T}_2\left(\mathcal{T}_1 (\cdot) \right)$. Thus, we can study invariance for basic transformations, and the results hold for a combination of these transformations. In this paper, we mainly consider similarity transformations.
\begin{definition}[Similarity Transformation]\label{def:trans}
	A similarity transformation is an arbitrary combination of the following transformations: (1) (Uniform) Scaling: the coordinate matrix is scaled by a constant, i.e., $\mathcal{T}(\mathbf{F}) = c \mathbf{F}$, where $c \neq 0$ is a constant. (2) Isometric transformation: any transformation that is isometric with respect to the metric $\mathcal{D}(\cdot,\cdot)$, i.e.,
	\begin{equation}\label{eq:iso}
		\left\{ \mathcal{T}(\cdot): \mathcal{D}\left(\mathbf{F}_{i,:},\mathbf{F}_{j,:}\right) = \mathcal{D}\left(\mathcal{T}(\mathbf{F})_{i,:},\mathcal{T}(\mathbf{F})_{j,:}\right),\forall \mathbf{F}, i,j \right\}.
	\end{equation}
	For the Euclidean distance, isometric transformations include rotation, translation, and reflection.
\end{definition}
Without proper designs, most message-passing GNNs and other neural networks for point clouds cannot guarantee transformation invariance for geometric data.

\section{Methodology}
In this section, we first revisit transformation invariance, and then introduce our proposed method and provide some discussions. We further provide extending our method to equivariance cases in Appendix~\ref{sec:equ}.
\subsection{Revisiting Transformation Invariance}\label{sec:visiting}
To investigate why typical neural network are not transformation invariant, we revisit transformation invariance of geometric data. We use GNNs as examples, but the analyses generalize to other neural networks following Eq.~\eqref{eq:geomnn}. 

We denote $\text{AGG}^{(l)}(\cdot)$ as an order-invariant aggregation function and $\text{COMBINE}^{(l)}(\cdot)$ as a combining function.
The message-passing framework of GNNs~\cite{gilmer2017neural} is formulated as:
\begin{equation}\label{eq:MPNN}
	\begin{split}
		\mathbf{m}^{(l)}_i = \text{AGG}^{(l)}\left(\left\{\mathbf{h}^{(l)}_j, \forall j \in \mathcal{N}(i) \right\} \right)\\ 
		\mathbf{h}^{(l+1)}_i = \sigma\left(\text{COMBINE}^{(l)}\left[ \mathbf{m}^{(l)}_i, \mathbf{h}^{(l)}_i\right] \right)
	\end{split},
\end{equation}
where $\mathbf{h}^{(l)}_i$ denotes the representation of point $v_i$ at the $l^{th}$ layer, $\mathbf{m}_i^{(l)}$ is the message vector for point $v_i$, and $\sigma(\cdot)$ is an activation function. We denote $\mathbf{H}^{(l)} = \left[ \mathbf{H}^{(l)}_1,...,\mathbf{H}^{(l)}_N \right]$ as the representation of all the points. The initial representation is 
\begin{equation}
	\mathbf{H}^{(0)} = \mathcal{P}\left(\mathbf{F}\right),
\end{equation} 
where $\mathcal{P}(\cdot)$ is the mapping function. The final representation is $\mathbf{H} = \mathbf{H}^{(L)}$, where $L$ is the number of layers. We easily have the following remark.
\begin{remark}\label{re:init}
	A GNN following Eq.~\eqref{eq:MPNN} is transformation-invariant if $\mathcal{P}(\cdot)$ is transformation-invariant.
\end{remark}
The remark can be proven by mathematical induction, i.e., if $\mathbf{H}^{(l)}$ is transformation-invariant, $\mathbf{H}^{(l+1)}$ is also transformation-invariant. Remark~\ref{re:init} shows that to empower GNNs to be transformation invariant, we simply need to ensure that the initial mapping function is transformation-invariant. However, the existing GNNs directly adopt the coordinates as the initial representations, i.e., $\mathcal{P}(\mathbf{F}) =  \mathbf{F} = \mathbf{H}^{(0)}$ , and thus cannot satisfy Remark~\ref{re:init}. It is natural to ask: can we have a principled method to obtain transformation-invariant initial representation from the coordinates? If so, we can realize transformation invariant neural networks without modifying the message-passing mechanism.

Manually designing heuristics is obviously one choice. For example, we can calculate the distances and angles of points with their nearest neighbors, i.e., the kNN method. However, vital information may be lost in the heuristics, leading to sub-optimal results. Ideally, we expect the mapping function $\mathcal{P}(\cdot)$ to be ``information lossless' so that $\mathbf{H}^{(0)}$ contains the same amount of information as $\mathbf{F}$.

For transformation-invariant geometric problems, useful information is encoded in the relative distance between points instead of the coordinates per se. Thus, if $\mathbf{H}^{(0)}$ can be distance-preserving, it is safe to say it is information lossless. We formulate the distance-preserving requirement as:
\begin{equation}\label{eq:reqdis1}
	\mathcal{D}(\mathbf{H}^{(0)}_{i,:},\mathbf{H}^{(0)}_{j,:}) = \mathcal{D}\left(\mathbf{F}_{i,:},\mathbf{F}_{j,:}\right), \forall i,j.
\end{equation}
Besides, the transformation-invariant requirement is formulated as
\begin{equation}\label{eq:reqinv}
	\mathcal{P} \left(\mathbf{F}\right) = \mathcal{P} \left(\mathcal{T}\left(\mathbf{F} \right)\right), \forall \; \mathbf{F},
\end{equation}
where $\mathcal{T}(\cdot)$ is any transformation in Definition~\ref{def:trans}.

However, there exists a conflict between Eq.~\eqref{eq:reqdis1} and Eq.~\eqref{eq:reqinv} for the scaling transformation. Specifically, when $\mathcal{T}(\mathbf{F}) = c \mathbf{F}$, the distance between points scales accordingly, i.e., $\mathcal{D}\left(c\mathbf{F}_{i,:},c\mathbf{F}_{j,:}\right) = c \mathcal{D}\left(\mathbf{F}_{i,:},\mathbf{F}_{j,:}\right)$, since the Euclidean distance has homogeneity of degree $1$. However, Eq.~\eqref{eq:reqinv} requires the new features to be invariant, i.e., $\mathbf{H}^{(0)}= \mathcal{P}\left(c\mathbf{F} \right) = \mathcal{P}\left(\mathbf{F} \right)$, and thus $\mathcal{D}(\mathbf{H}^{(0)}_{i,:},\mathbf{H}^{(0)}_{j,:}) $ is also invariant. Therefore, Eq.~\eqref{eq:reqdis1} cannot hold when $c \neq 1$. 

To solve that conflict, we relax Eq.~\eqref{eq:reqdis1} by adding an additional scaling term, i.e., assuming there exists a constant $c^\prime$ so that
\begin{equation}\label{eq:reqdis2}
	\mathcal{D}\left(\mathbf{H}^{(0)}_{i,:},\mathbf{H}^{(0)}_{j,:}\right) = c^\prime \mathcal{D}\left(\mathbf{F}_{i,:},\mathbf{F}_{j,:}\right), \forall i,j.
\end{equation}
In other words, the distance-preserving requirement is relaxed to not care for the absolute scale of the distance, but only preserve the relative ratios between different distances.

In summary, we require $\mathcal{P}(\cdot)$ to simultaneously satisfy Eqs.~\eqref{eq:reqinv}~\eqref{eq:reqdis2}. Then, using Remark~\ref{re:init}, we can adopt the initial point representation obtained by $\mathcal{P}(\cdot)$ to realize transformation-invariant neural networks. Notice that the solution to these two constraints is not unique. Therefore, we also require $\mathcal{P}(\cdot)$ to work in a deterministic way to ensure invariance. Next, we introduce our proposed plug-in to instantiate $\mathcal{P}(\cdot)$  that satisfies these requirements.

\subsection{The \model Method}\label{sec:model}
In this section, we present our proposed method based on the findings in Section~\ref{sec:visiting}. Specifically, we find that we can easily achieve the goal by slightly modifying multi-dimensional scaling (MDS), a classical dimensionality reduction technique~\cite{cox2008multidimensional}, as a plug-in for the existing neural networks. 

The core idea of MDS is to obtain distance-preserving features by an eigen-decomposition problem.  Specifically, we denote the distance matrix as $\mathbf{D}_{i,j} = \mathcal{D}\left(\mathbf{F}_{i,:},\mathbf{F}_{j,:}\right)$ and further construct a similarity matrix $\mathbf{S}\in\mathbb{R}^{N\times N}$ as follows:
\begin{equation}\label{eq:sim1}
	\mathbf{S}_{i,j} = -\frac{1}{2} \mathbf{D}_{i,j}^2 = -\frac{1}{2}\left(\mathcal{D}\left(\mathbf{F}_{i,:},\mathbf{F}_{j,:}\right)\right)^2.
\end{equation}
Then, we center the similarity matrix by
\begin{equation}\label{eq:sim2}
	\tilde{\mathbf{S}}_{i,j} = \mathbf{S}_{i,j} - \bar{\mathbf{S}}_{i,\cdot} - \bar{\mathbf{S}}_{\cdot,j} + \bar{\mathbf{S}}_{\cdot ,\cdot},
\end{equation}
where 
\begin{equation}
	\begin{split}
		\bar{\mathbf{S}}_{i,\cdot} = \frac{1}{N} \sum \nolimits_{k=1}^N \mathbf{S}_{i,k},\; \bar{\mathbf{S}}_{\cdot,j} = \frac{1}{N} \sum \nolimits_{l=1}^N \mathbf{S}_{l,j},\\ \;\bar{\mathbf{S}}_{\cdot,\cdot} = \frac{1}{N^2} \sum \nolimits_{k=1}^N \sum\nolimits_{l=1}^N \mathbf{S}_{k,l},
	\end{split}
\end{equation}
i.e., the average of the $i^{\text{th}}$ row, the average of the $j^{\text{th}}$ column, and the average of the matrix, respectively. We can combine Eqs.~\eqref{eq:sim1}~\eqref{eq:sim2} in an equivalent matrix form:
\begin{equation}\label{eq:sim3}
	\tilde{\mathbf{S}} = -\frac{1}{2} \left(\mathbf{I}_N - \frac{1}{N}\mathbf{1}_N \right) \left(\mathbf{D} \odot \mathbf{D}\right) \left(\mathbf{I}_N - \frac{1}{N}\mathbf{1}_N  \right),
\end{equation}
where $\mathbf{I}_N$ is a $N \times N$ identity matrix, $\mathbf{1}_N$ is a $N \times N$ matrix of ones, and $\odot$ is the Hadamard product. 

$\tilde{\mathbf{S}}$ is a $N\times N$ matrix containing the information of distances. To reduce the dimensionality, we calculate the eigen-decomposition of $\tilde{\mathbf{S}}$. We denote the eigenvalues of $\tilde{\mathbf{S}}$ sorted in descending order as a diagonal matrix $\mathbf{\Lambda}$, i.e., $\mathbf{\Lambda}_{1,1} \geq \mathbf{\Lambda}_{2,2} \geq ... \geq \mathbf{\Lambda}_{N,N}$ are eigenvalues, and $\mathbf{X}$ is a matrix of eigenvectors with $\mathbf{X}_{:,i}$ being the eigenvector associated with $\mathbf{\Lambda}_{i,i}$. The point representation is:
\begin{equation}\label{eq:MDS1}
	\tilde{\mathbf{H}}^{(0)} = \mathbf{X} \sqrt{\mathbf{\Lambda}}.
\end{equation}
However, the original MDS in Eq.~\eqref{eq:MDS1} can only satisfy Eq.~\eqref{eq:reqdis1} but not Eq.~\eqref{eq:reqdis2}. Thus, we slightly modify Eq.~\eqref{eq:MDS1} to a normalized form
\begin{equation}\label{eq:MDS2}
	\mathbf{H}^{(0)} = \mathbf{X} \sqrt{\frac{\mathbf{\Lambda}}{\mathbf{\Lambda}_{1,1}}}.
\end{equation}
From the properties of MDS, Eq.~\eqref{eq:MDS2} exactly produces our desired $\mathbf{H}^{(0)} = \mathcal{P}(\mathbf{F})$. We formalize the results as follows.

\begin{theorem}\label{thm:thm1}
	If $\mathcal{D}(\cdot,\cdot)$ is the Euclidean distance, the point representation obtained in Eq.~\eqref{eq:MDS2} satisfies Eq.~\eqref{eq:reqdis2}, i.e., there exists a constant $c^\prime$ so that
	\begin{equation}
		\mathcal{D}(\mathbf{H}^{(0)}_{i,:},\mathbf{H}^{(0)}_{j,:}) = c^\prime \mathcal{D}\left(\mathbf{F}_{i,:},\mathbf{F}_{j,:}\right), \forall i,j.
	\end{equation}
\end{theorem}
\begin{theorem}\label{thm:thm3}
	The representation obtained in Eq.~\eqref{eq:MDS2} satisfies Eq.~\eqref{eq:reqinv}, i.e., given the coordinate matrix $\mathbf{F}$ and any $\mathcal{T}(\cdot)$ in Definition~\ref{def:trans}, the point representation $\mathbf{H}^{(0)}$ is invariant.
\end{theorem}
The proofs are provided in Section~\ref{sec:proof} in the appendix.

We show our overall framework in Algorithm~\ref{alg:MDS}. We name our proposed method \model to highlight that it is \textbf{T}ransformation \textbf{inv}ariant. Since our method is a general plug-in with alterable neural network components, it is extremely simple to combine with the existing GNN models (see line~6 of Algorithm~\ref{alg:MDS}). In fact, we can use any neural network in Eq.~\eqref{eq:geomnn} as the backbone of \modelns, including non-graph-based neural networks for geometric data.

\begin{algorithm}[t]
	\caption{\modelns: A Transformation-Invariant Neural Network Plug-in}
	\label{alg:MDS}
	\begin{algorithmic}[1]
		\REQUIRE The coordinate matrix $\mathbf{F}$, the distance metric $\mathcal{D}\left( \cdot,\cdot\right)$, the adjacency matrix $\mathbf{A}$ \\
		\STATE Calculate $\mathbf{D}_{i,j} = \mathcal{D}\left(\mathbf{F}_{i,:},\mathbf{F}_{j,:}\right), \forall i,j$
		\STATE Calculate $\tilde{\mathbf{S}}$ using Eq.~\eqref{eq:sim3}
		\STATE Calculate the eigenvalues $\mathbf{\Lambda}$ and eigenvectors $\mathbf{X}$ of $\tilde{\mathbf{S}}$
		\STATE Calculate $\mathbf{H}^{(0)}$ using Eq.~\eqref{eq:MDS2}
		\STATE Input $\mathbf{H}^{(0)}$ into neural networks, e.g., GNN message-passings in Eq.~\eqref{eq:MPNN} or general neural network in Eq.~\eqref{eq:geomnn}
	\end{algorithmic}
\end{algorithm}

\subsection{Discussions}\label{sec:discuss}
\subsubsection{Uniqueness of eigenvectors}\label{sec:unique}
One caveat to notice is that the eigenvectors can have arbitrary signs, i.e., $u$ and $-u$ are eigenvectors with the same eigenvalue. To tackle this ambiguity, canonical approaches can be adopted to determine the sign~\cite{eigensign}, e.g., by letting the sum of all values be positive. However, there are potential failure cases (e.g., the sum of values is zero) and issues regarding directions for different points. Therefore, we take another approach to enumerate all $2^d$ possible eigenvectors as a new data augmentation method. Notice that $d$ is typically small for real-world cases, e.g.,less than $3$. Therefore, the enumeration will result in slightly but not too heavy of computational burdens. We provide more justification and empirical evidence for such an approach in Section~\ref{sec:exp:aug}.

Another potential issue is eigenvalue multiplicity, i.e., multiple eigenvectors have the same eigenvalue. In that case, obtaining unique eigenvectors is more challenging. Luckily, we do not find eigenvalue multiplicity for our tested real-world datasets, and leave handling the issue as future works.

\subsubsection{Time complexity} The extra computational cost of \model compared to base models mainly comes from the eigen-decomposition of $\tilde{\mathbf{S}}$. Since it is easy to see that the rank of $\tilde{\mathbf{S}}$ is the same as raw feature $\mathbf{F}$, $\tilde{\mathbf{S}}$ has at most $d$ non-zero eigenvalues, where $d$ denotes the dimensionality of $\mathbf{F}$. Therefore, we only need to calculate the top-$d$ eigen-decomposition of $\tilde{\mathbf{S}}$, which has a time complexity $O(N^2d)$, where $N$ is the number of points. Experimental results to empirically support the time complexity analysis are provided in Section~\ref{sec:expscala}.

\subsubsection{Extension} One may also wonder whether \model can be generalized to non-Euclidean problems, e.g., the distance metric $\mathcal{D}(\cdot,\cdot)$ is not Euclidean. 
In those cases, \model can be directly adopted while guaranteeing transformation invariance, but the distance-preserving guarantee may not hold. To preserve non-Euclidean distances, we may need  generalized MDS~\cite{agarwal2007generalized,bronstein2006generalized}, non-linear dimensionality reduction methods~\cite{roweis2000nonlinear,tenenbaum2000global}, or other more advanced methods. We leave such explorations as future works.

\section{Experiments}
In this section, we conduct experiments to verify our proposed method. Specifically, we aim to
answer the following questions:
\begin{itemize}[leftmargin = 0.5cm]
	\item \textbf{Q1}: Can \model guarantee invariance with respect to various kinds of transformations in Definition~\ref{def:trans}, such as translation, rotation, and scaling?
	\item \textbf{Q2}: Can \model easily combine with different neural network architectures for geometric data? 
	\item \textbf{Q3}: How does \model perform compared with other invariant and non-invariant models?
\end{itemize}
Notice that we do not aim to create new records in the leaderboard. Instead, we aim to provide a fresh perspective for the geometric deep learning problem and empirically verify its usefulness.

\subsection{Point Cloud Analysis}\label{sec:pointcloud}
The point cloud is one important type of geometric data. 
Since the shape of objects is invariant to transformations such as rotations, transformation-invariant models are vital to point cloud analysis. We adopt two tasks: object classification and object part segmentation.

\begin{table*}[t]
	\caption{The results of point cloud analysis on the test set. The object classification results are accuracy (\%) on the ModelNet40 dataset. The object part segmentation results are the mean per-class IoU (\%) on the ShapeNet dataset. Larger values indicate better results for both tasks. Best results are in bold and ``---'' means the result is not reported in the paper. 
	}
	\label{tab:classification}
	\centering
	\begin{tabular}{c l | ccc | ccc} \toprule
		\multicolumn{2}{c}{Task}  & \multicolumn{3}{|c|}{Object Classification} &  \multicolumn{3}{c}{Object Part Segmentation} \\ \midrule  
		\multicolumn{2}{c|}{Setting}       		                                & z/z     &  SO3/SO3  &  z/SO3  &  z/z    &  SO3/SO3 &  z/SO3 \\ \midrule
		\multirow{4}{*}{\tabincell{c}{Invariant\\baselines}}     & RIConv~\cite{zhang2019rotation}      & 86.5    &  86.4     &  86.4   &  ---      &  75.5    &  75.3  \\
		& ClusterNet~\cite{chen2019clusternet}  & 87.1    &  87.1     &  87.1   &  ---      &  ---       &  ---     \\
		& PR-invNet~\cite{yu2020deep} 			& 89.2    &  89.2     &  89.2   &  79.4   &  79.4    &  79.4  \\ 
		& RI-GCN~\cite{kim2020rotation}         & 89.5    &  \textbf{89.5} &  \textbf{89.5}   &  ---      &  77.3    &  77.2  \\ \midrule
		\multirow{2}{*}{Base models}            & PointNet~\cite{qi2017pointnet}       & 87.0    &  63.6     &  13.4   &  81.0   &  71.4    &  29.0  \\ 
		& DGCNN~\cite{wang2019dynamic}         & \textbf{92.2}    &  73.3     &  22.3   &  \textbf{82.0}   &  75.9    &  29.6  \\ \midrule
		\multirow{2}{*}{Our method}     & \modelns(PointNet)     			    & 86.5    &  86.5     &  86.2   &  80.9   &  80.0    &  80.0  \\
		& \modelns(DGCNN)        			    & 89.5    &  \textbf{89.5}     &  \textbf{89.5}   &  \textbf{82.0}   &  \textbf{82.1}    &  \textbf{82.0}  \\ \bottomrule
	\end{tabular}
\end{table*}

\begin{table*}[th]
	\caption{Showcases of point cloud classification for rotated inputs from ModelNet40. All methods adopt the z/SO3 setting.}\label{tab:class}
	\vspace{-0.5cm}
	\begin{center}
		\scalebox{0.95}{
			\renewcommand{\arraystretch}{1.2}
			\setlength\tabcolsep{6pt}
			\begin{tabular}{l c c c c}
				Method & \includegraphics[width=0.15\textwidth] {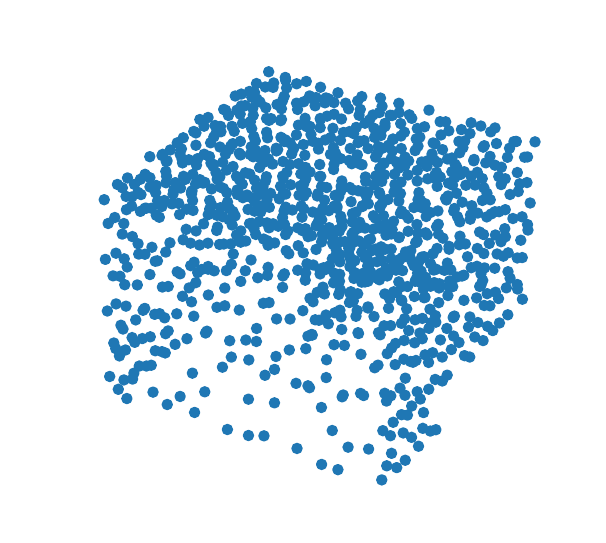} & \includegraphics[width=0.15\textwidth] {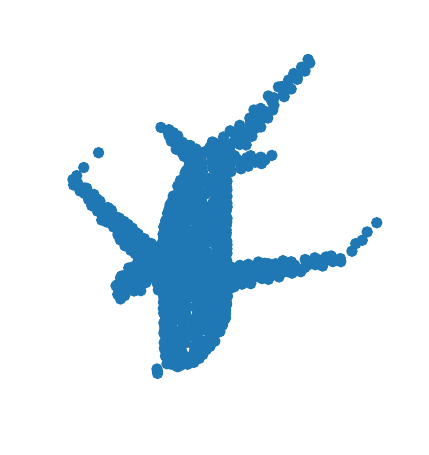} &   \includegraphics[width=0.15\textwidth] {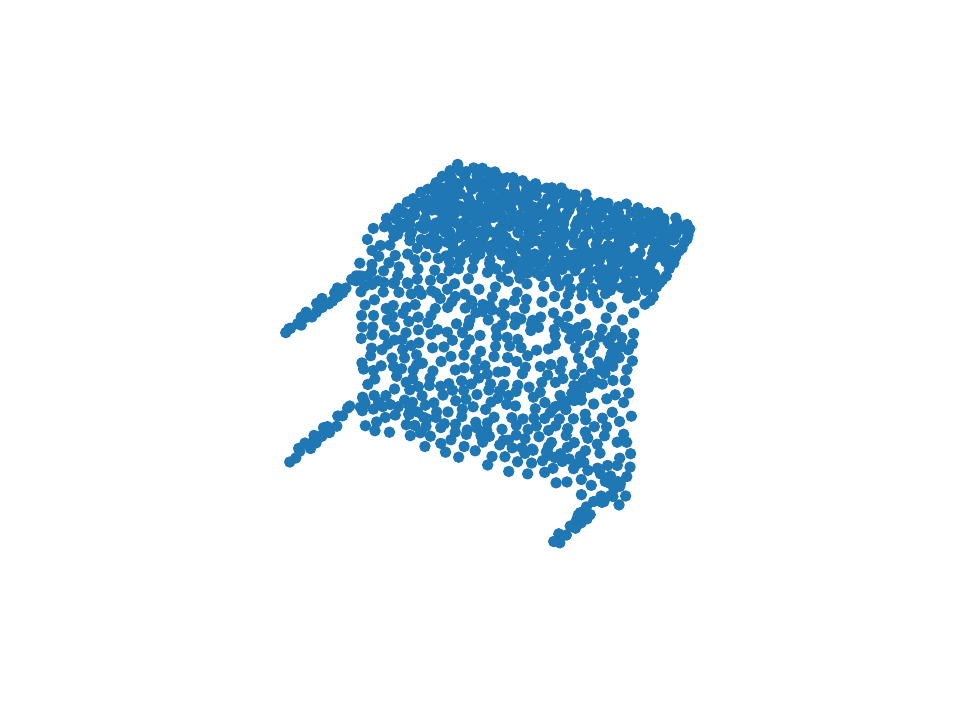} &  \includegraphics[width=0.18\textwidth] {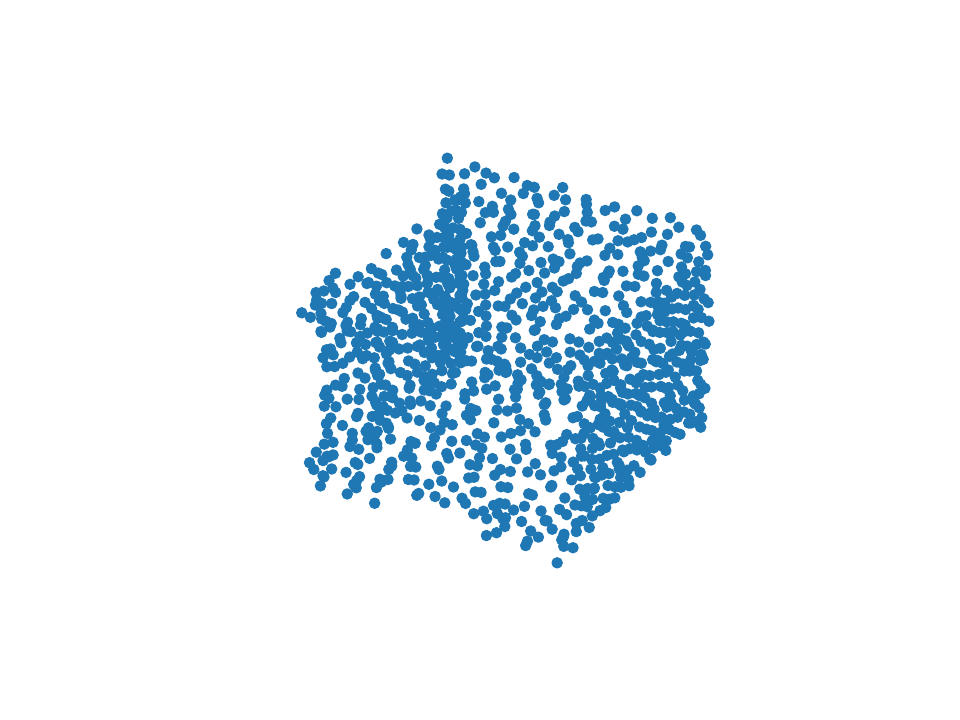}
				\\ \hline \hline 
				Ground Truth                   & \textbf{bookshelf}  & \textbf{airplane}  & \textbf{chair}   & \textbf{night-stand}   \\ \hline
				PointNet~\cite{qi2017pointnet} & car                 & desk               & stairs           & dresser                \\ \hline 
				DGCNN~\cite{wang2019dynamic}   & sofa                & \textbf{airplane}  & stairs           & glass-box              \\ \hline 
				\modelns(PointNet)             & \textbf{bookshelf}  & \textbf{airplane}  & \textbf{chair}   & \textbf{night-stand}   \\ \hline
				\modelns(DGCNN)                & \textbf{bookshelf}  & \textbf{airplane}  & \textbf{chair}   & \textbf{night-stand}   \\ \hline
			\end{tabular}
		}
	\end{center}
	\vspace{-0.25cm}
\end{table*}

\subsubsection{Object Classification}\label{sec:pointcloudclassification}
For object classification, we adopt ModelNet40~\cite{wu20153d}, a point cloud dataset containing 40 categories of CAD objects such as airplanes, cars, tables, etc. Each object is represented by 1,024 points with the 3-D coordinate of points. The task is to predict the categories of point clouds, i.e., a 40-class graph classification problem, since each point cloud can be considered a graph. We use a pre-processed dataset suggested in~\cite{qi2017pointnet}, containing 9,843 objects for training and 2,468 objects for testing. Following~\cite{zhang2019rotation,chen2019clusternet}, we adopt the following settings with different rotations. Notice that we only adopt rotation transformations here following the literature, while more transformations are adopted in combinatorial optimization problems in Section~\ref{sec:TSP}.
\begin{itemize}[leftmargin = 0.5cm]
	\item z/z: both training and testing data are augmented with rotations about the z-axis (the gravity axis).
	\item SO3/SO3: both training and testing data are augmented with arbitrary SO3 (3-D rotations).
	\item z/SO3: training data is augmented with z-axis rotations, and testing data is augmented with SO3.
\end{itemize}
The first setting is standard but less challenging since the data is aligned to known axes. The other two cases focus more on models' ability to handle rotations. The z/SO3 setting is most challenging since it requires models to generalize well to different rotations without seeing these rotations in the training data.  All results are measured by accuracy, i.e., how many percentages of point clouds are correctly classified. 

For baselines, we adopt both non-rotation-invariant models and rotation-invariant models. For the former, we adopt PointNet~\cite{qi2017pointnet} and DGCNN~\cite{wang2019dynamic}, which are widely adopted neural networks for geometric data. For rotation-invariant models, we adopt RIConv~\cite{zhang2019rotation}, ClusterNet~\cite{chen2019clusternet}, PR-invNet~\cite{yu2020deep}, and RI-GCN~\cite{kim2020rotation}, which are recently proposed rotation-invariant models. Notice that permutation-equivariant models cannot be simply applied here and thus are not compared, as in all previous works.

For our proposed method, we adopt PointNet and DGCNN, which are representative and effective neural networks but cannot maintain transformation invariance, as the backbone, i.e., using PointNet or DGCNN as line 5 of Algorithm~\ref{alg:MDS}, denoted as \modelns(PointNet) and \modelns(DGCNN), respectively. Notice that except for the initial representations, we keep other settings such as architectures and hyper-parameters unchanged (please refer to Appendix~\ref{sec:hp} for the exact settings). Thus, all changes in the performance can be attributed to the differences in initial representations. Besides, we would also emphasize that all the baselines are specially designed for point clouds, while our proposed \model is general and does not utilize any special property of point clouds. 

The results are shown in Table~\ref{tab:classification} and showcases are provided in Table~\ref{tab:class}. We make the following observations. 

Base models such as PointNet and DGCNN show promising results in the standard z/z setting, demonstrating their effectiveness in extracting useful information from point clouds. However, when generalizing to rotations, the performance drops significantly. In the most challenging z/SO3 setting, both PointNet and DGCNN perform miserably. Such results are consistent with the literature since PointNet and DGCNN do not consider rotations. 

Both \modelns(PointNet) and \modelns(DGCNN) perform equally well on all three settings, indicating \model can effectively handle  rotation transformations of point clouds. The results provide empirical evidence that \model is strictly transformation-invariant. 

The results in the SO3/SO3 setting show that adopting data augmentation in the training phase alleviates fairly the problem caused by rotations. For example, the performance of PointNet and DGCNN indeed improves when the training data is augmented with rotations. However, the performance gap between z/z and z/SO3 is still considerable. Since the number of possible 3D rotations is infinite, it is infeasible to enumerate all rotated objects.  

In the standard z/z setting, the results of \model closely match the corresponding backbone, i.e., \modelns(PointNet) as of PointNet and \modelns(DGCNN) as of DGCNN. The results demonstrate that \model does not lose useful information compared to using the coordinates, verifying our findings of distance-preserving. Notice that we use the identical hyper-parameters as in the original backbones. 

Our proposed method outperforms RIConv, ClusterNet and PR-invNet, and closely matches RI-GCN. Notice that these methods are specifically designed to be rotation-invariant, while \model is a simple and general method that does not depend on specific neural architectures. In other words, \model could be extended more easily to new advancements for geometric data, such as novel neural network architectures. 

\subsubsection{Object Part Segmentation}
For object part segmentation, we adopt ShapeNet~\cite{shapenet2015} that contains 16,880 CAD objects of 16 categories. Each object is represented by 2,048 points and has an annotation of 2 to 6 parts, adding up to 50 different parts in total. The task is to predict which part each point belongs to, i.e., a 50-class node classification problem since each point corresponds to one node. We follow the standard dataset splits with 14,006 objects for training and 2,874 objects for testing. Other settings such as transformations and baselines are the same as object classification in Section~\ref{sec:pointcloudclassification}.

\begin{table*}[t]
	\caption{The results of the travelling salesman problem (TSP) measured by the length of the output route (lower is better). The numbers in parentheses indicate the performance gap with respect to the specialized solver LHK3. }\label{tab:TSP}
	\begin{tabular}{l|l |P{1.05cm}P{0.85cm}P{1.2cm}P{1.05cm}P{1.1cm}P{0.95cm} P{1.2cm}P{1.0cm}P{1.2cm}P{0.95cm}} \toprule
		Size        & Method &  None & $_\text{(gap)}$ &  Translation  & $_\text{(gap)}$ &  Rotation & $_\text{(gap)}$ & Reflection & $_\text{(gap)}$  & Scaling\footnotemark & $_\text{(gap)}$  \\ \midrule
		\multirow{3}{*}{TSP-20}	
		& LKH3           &  3.83$\pm$0.01 &        & 3.83$\pm$0.01  &          & 3.83$\pm$0.01 &        & 3.83$\pm$0.01 &         &  3.83$\pm$0.01 &    \\
		& GAT            &  3.88$\pm$0.01 &(1.3\%) & 10.45$\pm$0.02 &(172.8\%) & 3.89$\pm$0.01 &(1.6\%) & 4.60$\pm$0.02 &(20.1\%) &  10.16$\pm$0.02& (165.4\%) \\
		& \model         &  3.87$\pm$0.01 &(1.1\%) & 3.87$\pm$0.01  &(1.1\%)   & 3.87$\pm$0.01 &(1.1\%) & 3.87$\pm$0.01 &(1.1\%)  &  3.87$\pm$0.01& (1.1\%) \\ \midrule
		\multirow{3}{*}{TSP-50}	
		& LKH3           &  5.69$\pm$0.01 &        & 5.69$\pm$0.01  &          & 5.69$\pm$0.01 &        & 5.69$\pm$0.01 &         &  5.69$\pm$0.01 &    \\
		& GAT            &  5.96$\pm$0.01 &(4.7\%) & 26.07$\pm$0.04 &(358.0\%) & 6.02$\pm$0.01 &(5.8\%) & 7.93$\pm$0.04 &(39.3\%) &  24.96$\pm$0.03& (338.6\%) \\
		& \model         &  5.98$\pm$0.01 &(5.1\%) & 5.98$\pm$0.01  &(5.1\%)   & 5.98$\pm$0.01 &(5.1\%) & 5.98$\pm$0.01 &(5.12\%) &  5.98$\pm$0.01 & (5.1\%) \\ \midrule
		\multirow{3}{*}{TSP-100}
		& LKH3           &  7.76$\pm$0.01 &        & 7.76$\pm$0.01  &          & 7.76$\pm$0.01 &        & 7.76$\pm$0.01 &         &  7.76$\pm$0.01 &    \\
		& GAT            &  8.49$\pm$0.01 &(9.4\%) & 52.15$\pm$0.06 &(571.7\%) & 8.66$\pm$0.01 &(11.5\%)& 21.60$\pm$0.09&(178.2\%)&  52.83$\pm$0.04& (580.7\%) \\
		& \model         &  8.52$\pm$0.01 &(9.8\%) & 8.52$\pm$0.01  &(9.8\%)   & 8.52$\pm$0.01 &(9.8\%) & 8.52$\pm$0.01 &(9.8\%)  &  8.52$\pm$0.01 & (9.8\%) \\ \bottomrule
	\end{tabular}
\end{table*}

\begin{table*}[t]
	\caption{The results of the capacitated vehicle routing problem (CVRP) measured by the length of the output route (lower is better). The numbers in parentheses indicate the performance gap with respect to the specialized solver LHK3.}\label{tab:CVRP}
	\begin{tabular}{P{1.3cm}|p{1.15cm} |P{1.2cm}P{0.8cm}P{1.3cm}P{0.95cm}P{1.15cm}P{0.8cm} P{1.15cm}P{0.95cm}P{1.35cm}P{0.95cm}} \toprule
		Size        & Method & None & $_\text{(gap)}$ & Translation & $_\text{(gap)}$ &  Rotation & $_\text{(gap)}$ & Reflection & $_\text{(gap)}$ & Scaling\footnotemark[2] & $_\text{(gap)}$ \\ \midrule
		\multirow{3}{*}{CVRP-20}   
		& LKH3        & 6.13$\pm$0.02 &        & 6.13$\pm$0.02 &         & 6.13$\pm$0.02 &         & 6.13$\pm$0.02 &         & 6.13$\pm$0.02 &         \\
		& GAT         & 6.55$\pm$0.02 &(6.8\%) &20.86$\pm$0.08&(240.5\%) & 6.57$\pm$0.02 & (7.2\%) & 7.84$\pm$0.03 &(28.0\%) & 20.23$\pm$0.06&(230.3\%)\\
		& \model      & 6.56$\pm$0.02 &(7.1\%) & 6.56$\pm$0.02 &(7.1\%)  & 6.56$\pm$0.02 & (7.1\%) & 6.56$\pm$0.02 &(7.1\%)  & 6.56$\pm$0.02 &(7.1\%)  \\ \midrule
		\multirow{3}{*}{CVRP-50}   
		& LKH3        & 10.36$\pm$0.02&        & 10.36$\pm$0.02&         & 10.36$\pm$0.02&         & 10.36$\pm$0.02&         & 10.36$\pm$0.02&         \\
		& GAT         & 11.31$\pm$0.03&(9.1\%) & 52.09$\pm$0.18&(402.8\%)& 11.37$\pm$0.03& (9.8\%) & 15.56$\pm$0.07&(50.2\%) & 48.14$\pm$0.12&(364.7\%)\\
		& \model      & 11.38$\pm$0.03&(9.8\%) & 11.38$\pm$0.03&(9.8\%)  & 11.38$\pm$0.03& (9.8\%) & 11.38$\pm$0.03&(9.8\%)  & 11.38$\pm$0.03&(9.8\%)  \\ \midrule
		\multirow{3}{*}{CVRP-100}  
		& LKH3        & 15.61$\pm$0.04&        & 15.61$\pm$0.04&         & 15.61$\pm$0.04&         & 15.61$\pm$0.04&         & 15.61$\pm$0.04&         \\
		& GAT         & 17.73$\pm$0.04&(13.6\%)&104.23$\pm$0.35&(567.8\%)& 17.87$\pm$0.04&(14.5\%) & 62.95$\pm$0.50&(303.3\%)&100.95$\pm$0.24& (546.6\%)\\
		& \model      & 17.31$\pm$0.04&(10.9\%)& 17.31$\pm$0.04&(10.9\%) & 17.31$\pm$0.04& (10.9\%)& 17.31$\pm$0.04&(10.9\%) & 17.31$\pm$0.04& (10.9\%) \\ \bottomrule
	\end{tabular}
\end{table*}
We report the results in Table~\ref{tab:classification} and provide some showcases in Figure~\ref{fig:segmentation}.
Our proposed method \model manages to beat all comparing methods for object part segmentation in the SO3/SO3 and z/SO3 setting. The results reconfirm that \model is highly capable of handling rotation transformations of point clouds. Though other invariant baselines are not affected by rotations, they fail to be as expressive as our proposed method. Besides, owing to the simplicity and general applicability of \modelns, we expect the performance to improve further when adopting more powerful neural networks.  

\begin{figure*}[ht]
	\captionsetup[subfigure]{font=small,labelfont=small}
	\centering
	\subcaptionbox{Ground truth}
	{\includegraphics[width=0.19\linewidth]{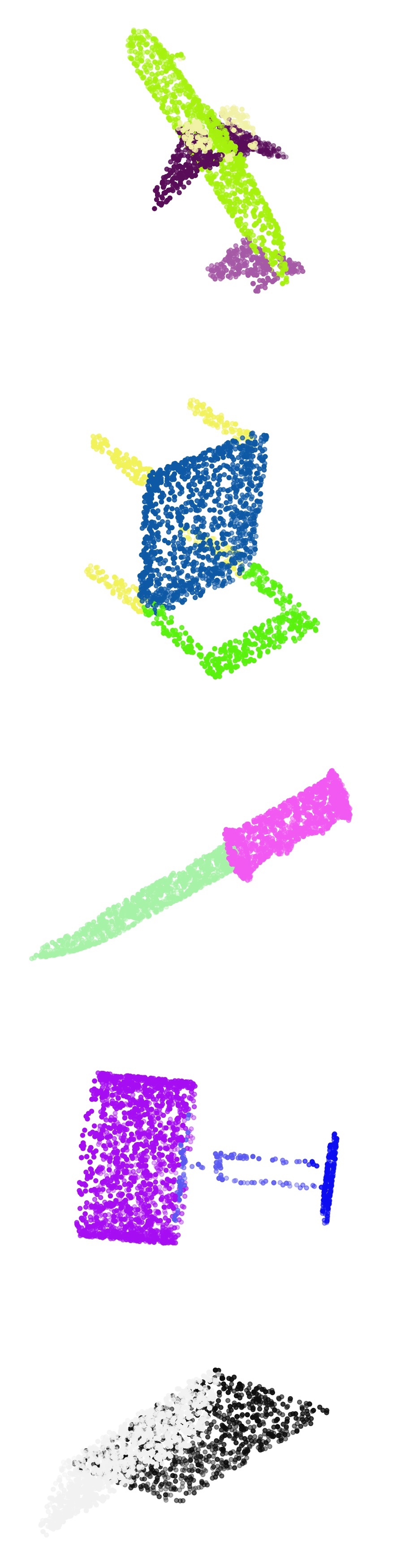}}
	\subcaptionbox{PointNet}
	{\includegraphics[width=0.19\linewidth]{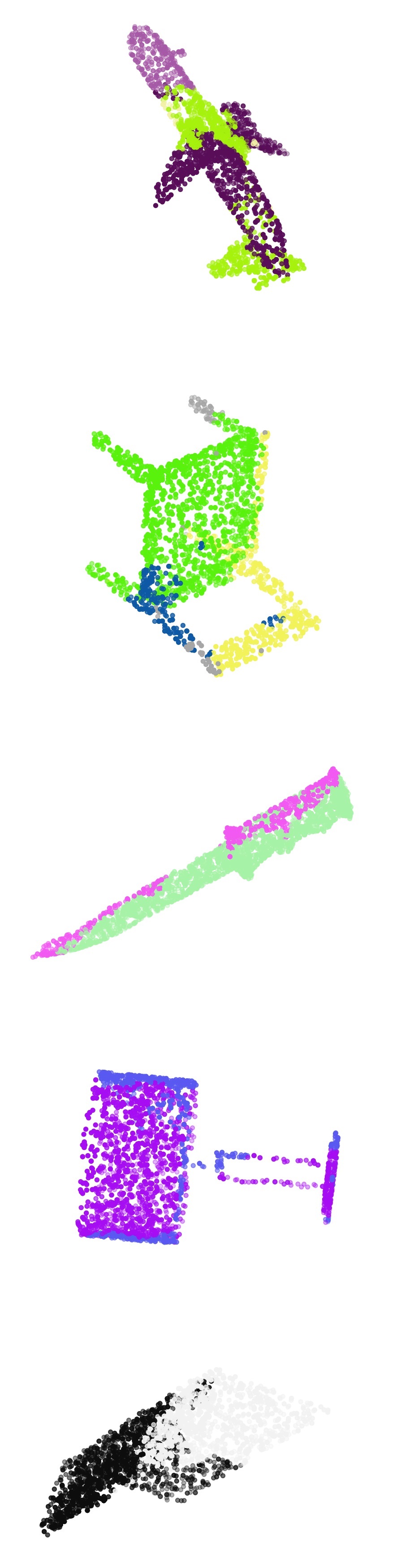}}
	\subcaptionbox{DGCNN}
	{\includegraphics[width=0.19\linewidth]{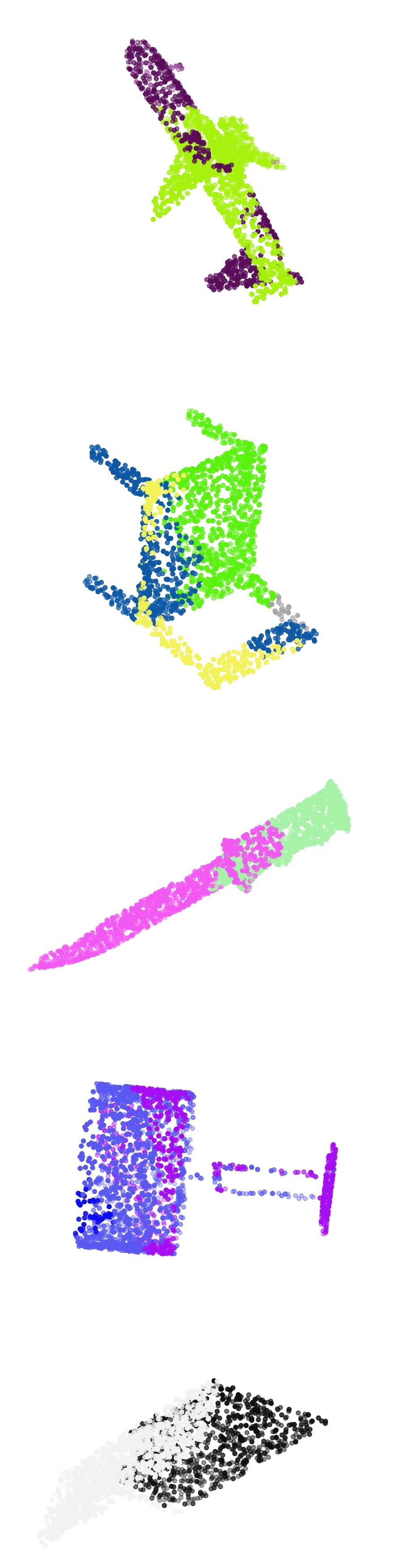}}
	\subcaptionbox{\modelns(PointNet)}
	{\includegraphics[width=0.19\linewidth]{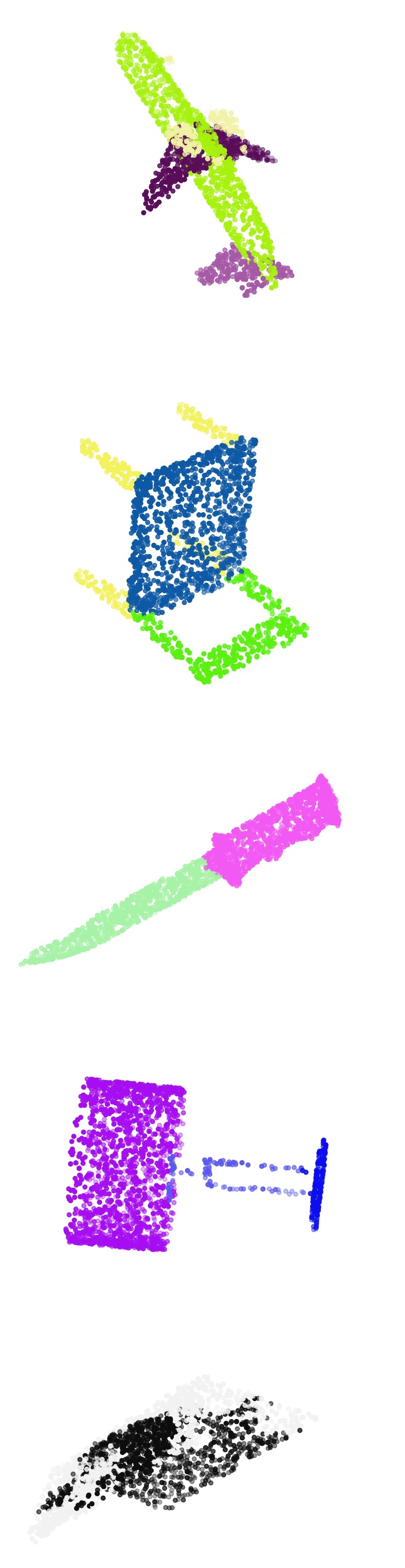}}
	\subcaptionbox{\modelns(DGCNN)}
	{\includegraphics[width=0.19\linewidth]{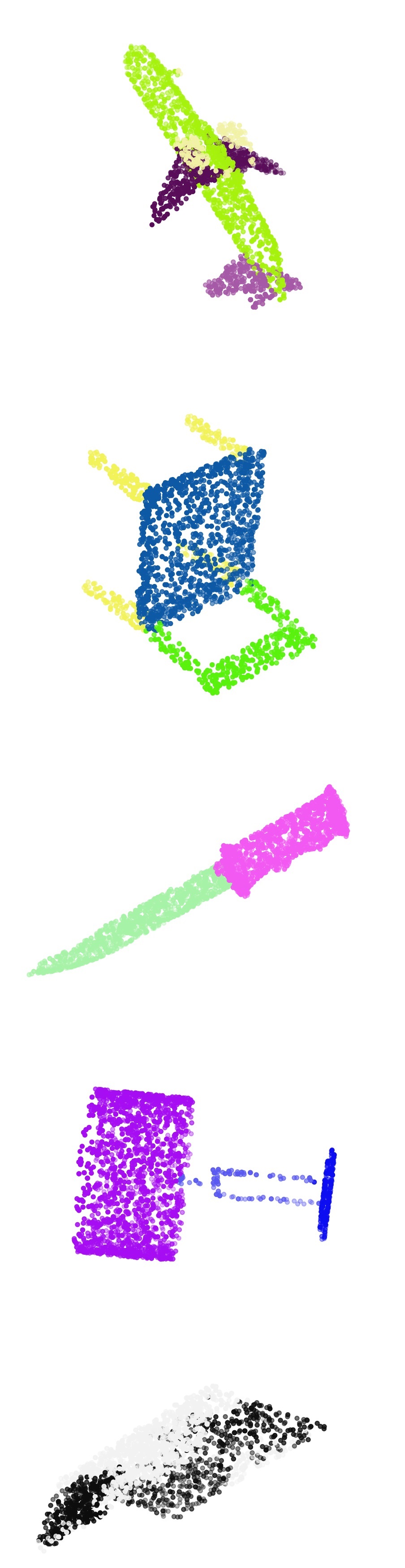}}
	\caption{Showcases of point cloud part segmentation for rotated inputs from ShapeNet. All methods adopt the z/SO3 setting.}
	\label{fig:segmentation} 
\end{figure*}
In summary, the results of object classification and object part segmentation clearly demonstrate the effectiveness of \model in handling rotation transformations of geometric data, outperforming or matching baselines. 

\subsection{Combinatorial Optimization}
Geometric data combined with simple objectives and constraints form challenging combinatorial optimization problems. Using neural networks to solve combinarotial optimization has a long history~\cite{smith1999neural} with many important applications~\cite{bengio2020machine}. Therefore, we test the effectiveness of \model in two such tasks: the travelling salesman problem and the capacitated vehicle routing problem.
\subsubsection{Travelling Salesman Problem}\label{sec:TSP}
Travelling salesman problem (TSP) is a well-known NP-hard problem~\cite{papadimitriou1977euclidean} with many practical applications such as logistics and scheduling~\cite{lenstra1975some}. 
Given a set of points and the distances between points, TSP aims to find the shortest possible route that visits each point exactly once and returns to the origin point. 
Due to the difficulty in finding optimal solutions, learning-based methods have been adopted to facilitate TSP solvers~\cite{bengio2020machine}. 

We follow \cite{kool2018attention} for the experimental setting.  We consider three cases: TSP-20, TSP-50, and TSP-100, containing $N=20$, $N=50$, and $N=100$ points per data instance, respectively. For each case, we randomly generate 100,000 instances for training and 10,000 instances for testing. All the points have a random 2-D coordinate in the range $[0,1]$. The distance between points is the Euclidean distance. We adopt the following settings with different transformations of the coordinates:
\begin{itemize}[leftmargin = 0.5cm]
	\item \textbf{None}: The coordinate of the points is kept the same as the raw coordinates.  
	\item \textbf{Translation}: We add a random constant to the coordinate of the points, i.e., $\mathcal{T}(\mathbf{F}) = \mathbf{F} + c$, where the constant $c$ is drawn uniformly from $[-100,100]$.
	\item \textbf{Rotation}: The coordinate of the points is randomly rotated with respect to the centroid, i.e.,  $\mathcal{T}(\mathbf{F})_{i,:} = \bar{\mathbf{F}} + \left(\mathbf{F}_{i,:} - \bar{\mathbf{F}} \right) \mathbf{R}$, where $\bar{\mathbf{F}} = \frac{1}{N}\sum_{i=1}^N \mathbf{F}_{i,:}$ is the centroid and $\mathbf{R} \in \text{SO}2$ is a random 2-D rotation matrix.
	\item \textbf{Reflection}: The coordinate is reflected with respect to the $x$-axis, i.e., $\mathcal{T}(\mathbf{F})_{i,j} = \mathbf{F}_{i,j} (-1)^j$.
	\item \textbf{Scaling}: The coordinate of the points is uniformly scaled, i.e., $\mathcal{T}(\mathbf{F}) = c \mathbf{F}$, where $c$ is a random number drawn uniformly from $(0,100]$.
\end{itemize} 
It is easy to that the optimal routes of TSP should be invariant to the above transformations. 

For the baseline and our proposed method, we adopt GAT~\cite{velivckovic2018graph} as the GNN backbone, which has been shown to outperform various other GNNs~\cite{kool2018attention}. The detailed settings, e.g., architectures, hyper-parameters, and decoder structures, are kept the same as in the original paper (please refer to Appendix~\ref{sec:hp} for details). Besides, we adopt LKH3~\cite{helsgaun2017lkh}, a specialized solver for TSP, as a reference line. Notice that, similar to Section~\ref{sec:pointcloud}, we do not alter our model to consider the special characteristics of TSP. Instead, we want to demonstrate the effectiveness of \model in handling various transformations.

We show the results in Table~\ref{tab:TSP}. \model achieves identical results across different transformations, including translation, rotation, reflection, and scaling, verifying Theorem~\ref{thm:thm3} that our proposed method is strictly transformation invariant to all the transformations in Definition~\ref{def:trans}. Notice that though we study these transformations independently, the compositionality of transformation invariance guarantees that \model can also handle an arbitrary combination of these transformations. 

Though GAT achieves similar results as \model in the \textbf{None} setting, it performs poorly when transformations are applied. The results show that GAT is most sensitive to translation and scaling since the range of input features differs significantly compared to the coordinates seen during training. The results are consistent with the literature~\cite{tang2020towards}. In contrast, our \model model does not suffer this issue and greatly outperforms GAT. 

There are still gaps between learning-based methods and the specialized solver LHK3, and the gap grows larger as the problem size grows. These results indicate that novel neural networks and training strategies are still needed to advance further the study of learning to solve combinatorial optimizations. Note that though the existing dedicated solvers may lead to better results for certain problems at the current stage, we believe it is important to continue the study of machine learning based combinatorial optimization methods with potential benefits including scalability to extremely large-scale instances, generalization to new problems, inspiring new solvers, etc~\cite{smith1999neural,bengio2020machine}. Since \model is simple and general, we expect \model to be easily extended to these yet-to-come methods. Future studies may only concern the \textbf{None} case to design more powerful architectures and let \model handle the transformation invariance problem.

We also plot the output route of GAT and \model for one example instance of TSP-20 in Figure~\ref{fig:TSP}. The figure clearly demonstrates the importance of maintaining transformation invariance and the efficacy of \modelns.

\subsubsection{Capacitated Vehicle Routing Problem}\label{sec:cvrp}
Capacitated vehicle routing problem (CVRP)~\cite{toth2014vehicle} is a generalization to TSP with more practical usages. Given a set of points and the distances between points, instead of finding one shortest route as in TSP, we need to construct multiple routes starting and ending from a central depot. The goal is to minimize the length of all the routes while meeting the demand of each point. Besides, the total demand of points in each route is capacitated, i.e., corresponding to constraints in real delivery problems that our ``vehicles'' have a limited capacity. CVRP is also known to be an NP-hard problem~\cite{toth2014vehicle}. 

Similar to Section~\ref{sec:TSP}, we follow \cite{kool2018attention} to set up the experimental setting for CVRP. Specifically, we adopt three cases: CVRP-20, CVRP-50, and CVRP-100 containing 20, 50, and 100 points per instance, respectively. The coordinates of points are in the range $[0,1]$, and the metric is the Euclidean distance. For each case, we generate 100,000 instances for training and 10,000 instances for testing. We adopt the same five transformations as in TSP, i.e., \textbf{None}, \textbf{Translation}, \textbf{Rotation}, \textbf{Reflection}, and \textbf{Scaling}. The baselines and experimental settings are also the same as in Section~\ref{sec:TSP}. For more details, please refer to Appendix~\ref{sec:hp}.

We report the results in Table~\ref{tab:CVRP} and provide a showcase in Figure~\ref{fig:cvrp}. 
The results show similar trends as in Table~\ref{tab:TSP} for TSP. Concretely, GAT performs reasonably well in the \textbf{None} setting but fails to generalize to different transformations. For some transformations like translation and scaling, the results of GAT even become intolerable.  On the other hand, thanks to the transformation invariance property of \modelns, it is able to handle different transformations with zero performance drop. The results demonstrate again that \model is an effective and general solution towards transformation invariant combinatorial problems.    

\begin{figure*}[ht]
	\centering
	\subcaptionbox{The results of GAT.}
	{\includegraphics[width=0.99\linewidth]{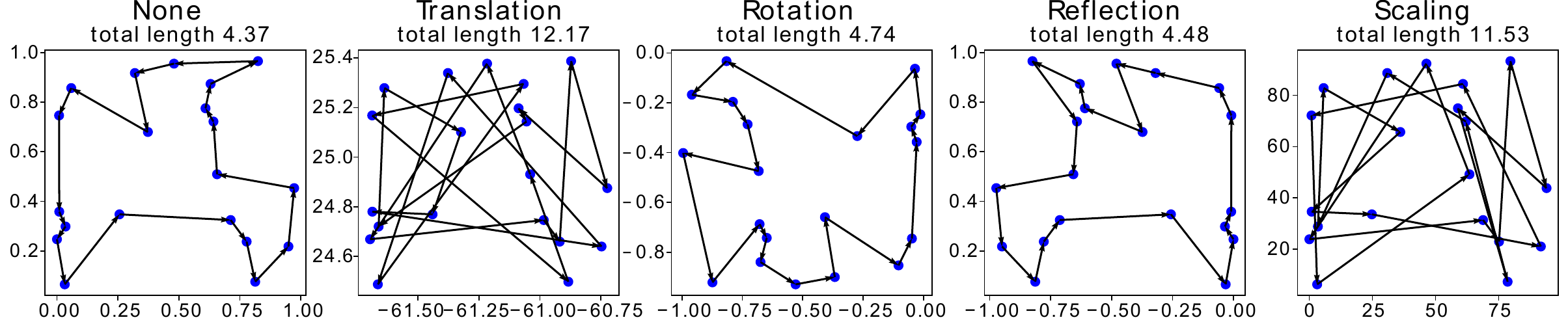}}
	\subcaptionbox{The results of \modelns.}
	{\includegraphics[width=0.99\linewidth]{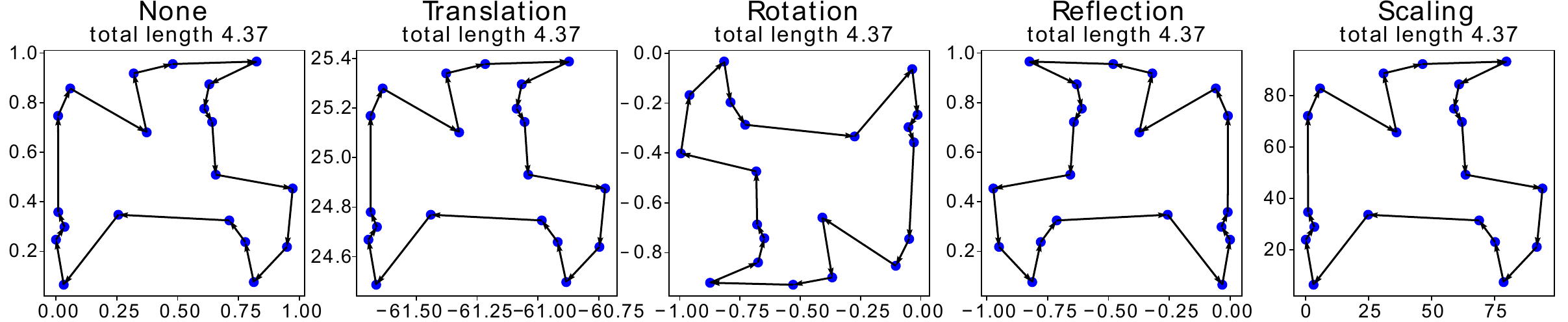}}
	\caption{The output route of (a) GAT, and (b) \model for one example instance of TSP-20 after different transformations.}
	\label{fig:TSP} 
\end{figure*}

\begin{figure*}[ht]
	\centering
	\subcaptionbox{The results of GAT.}
	{\includegraphics[width=0.99\linewidth]{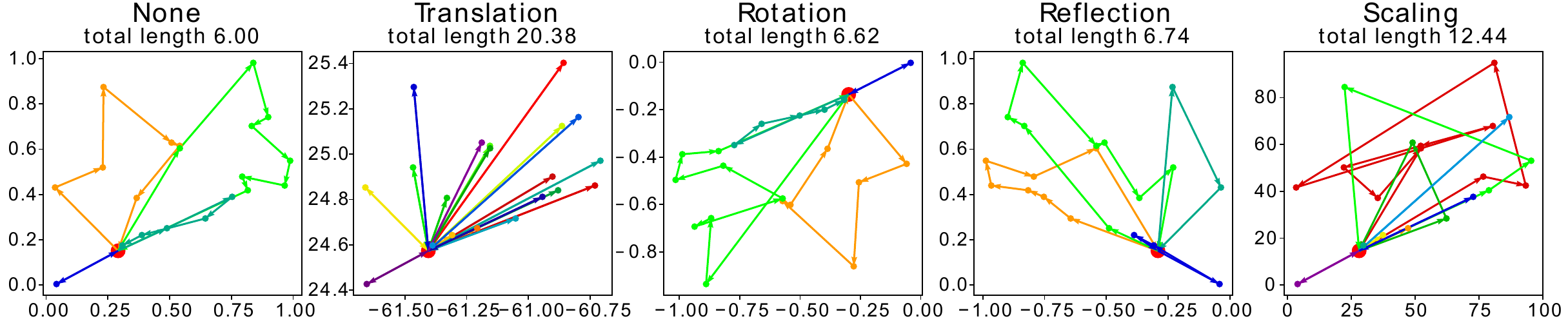}}
	\subcaptionbox{The results of \modelns.}
	{\includegraphics[width=0.99\linewidth]{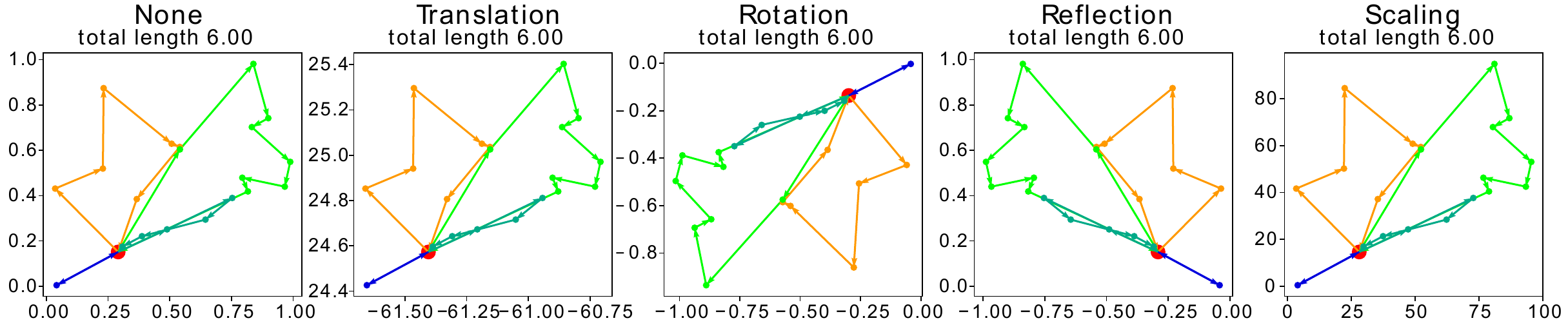}}
	\caption{The output route of (a) GAT, and (b) \model for one example instance of CVRP-20 after different transformations.}
	\label{fig:cvrp} 
\end{figure*}

\subsection{Analysis}
\subsubsection{Scalability}\label{sec:expscala}
To empirically analyze the scalability of our proposed method, we record the running time of calculating the initial point representations while varying the number of points. The average results of 10 runs are reported in Figure~\ref{fig:scala}. We also fit a linear regression curve after applying log transformation on both axes and report the fitting statistics. 
\begin{figure}[t]
	\centering
	\includegraphics[width=5cm]{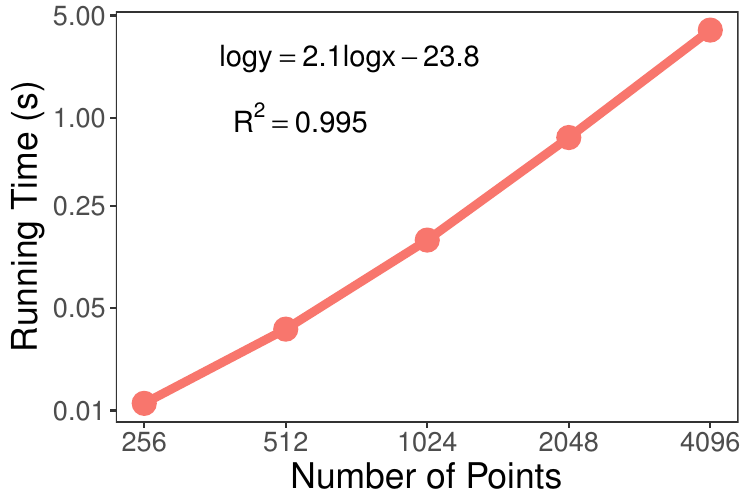}\\
	\caption{The running time of calculating the initial point representations with respect to the number of points.}\label{fig:scala}
\end{figure}
The results show that the running time grows quadratically with respect to the number of points, which is consistent with our analysis in Section~\ref{sec:model}. Besides, for a data instance with 2,048 points, the running time is less than 1 second. Notice that the initial point representations only need to be calculated once and can be pre-computed, while the optimization of backbone neural networks usually needs dozens or hundreds of epochs. Therefore, \model does not incur high extra computational costs.
\footnotetext{Since the length of routes is proportional to scaling, we normalize the results in the scaling setting to be consistent with other settings, i.e., when $\mathcal{T}(\mathbf{F}) = c \mathbf{F}$, the length of output routes is scaled by $\frac{1}{c}$.}

\subsubsection{Comparison with kNN and Data Whitening}
A straightforward heuristic to obtain transformation-invariant features is $k$-NN, i.e., calculating the distance of each point with its $k$ nearest neighbors. Besides, data whitening is a classical data pre-processing method by decorrelating different dimensions of the input data. From its properties, data whitening can also ensure transformation-invariance (but not distance-preserving). 
Next, we empirically compare our proposed method with $k$-NN and data whitening. Specifically, we choose two $k$ values for $k$-NN: $k=3$, which has the same dimensionality as raw features and our proposed method, and $k=10$, which contains more flexibility. For data whitening, we adopt the PCA-whitening. We report the results for the point cloud classification task in Table~\ref{tab:whitening}, while other tasks indicate similar results. The results show that, though $k$-NN and data whitening can maintain rotation-invariance, their performance is much lower than our proposed method.

\begin{table}[h]
	\caption{The results of comparing with kNN and data whitening for point cloud classification on the ModelNet40 dataset. The best results for each backbone are in bold.}
	\label{tab:whitening}
	\centering
	\begin{tabular}{l | ccc } \toprule
		Method                                      & z/z     &  SO3/SO3  &  z/SO3   \\ \midrule
		PointNet($k$-NN,$k$=3)                           & 24.9    &  24.9     &  24.9    \\ 
		PointNet($k$-NN,$k$=10)                          & 29.9    &  29.9     &  29.9    \\ 
		PointNet(Whitening)                         & 81.9    &  81.9     &  81.9    \\ 
		\modelns(PointNet)     			            & \textbf{86.5}    &  \textbf{86.5}     &  \textbf{86.2}    \\ \midrule
		DGCNN($k$-NN,$k$=3)                              & 29.9    &  29.9     &  29.9    \\ 
		DGCNN($k$-NN,$k$=3)                              & 36.1    &  36.1     &  36.1    \\ 
		DGCNN(Whitening)                            & 85.8    &  85.8     &  85.8    \\ 
		\modelns(DGCNN)        			            & \textbf{89.5}    &  \textbf{89.5}     &  \textbf{89.5}    \\ \bottomrule
	\end{tabular}
\end{table}
\begin{table}[t]
	\caption{The results of enumerating signs of eigenvectors as data augmentation and using a unique sign by the canonical approach for point cloud classification task.}
	\label{tab:cano}
	\centering
	\begin{tabular}{l | ccc } \toprule
		Method                                      & z/z     &  SO3/SO3  &  z/SO3   \\ \midrule
		\modelns(PointNet)-F						& 85.4    &  85.2     &  85.4    \\ 
		\modelns(PointNet)     			            & 86.5    &  86.5     &  86.2    \\ \midrule
		\modelns(DGCNN)-F        			        & 88.3    &  89.2     &  88.5    \\
		\modelns(DGCNN)        			            & 89.5    &  89.5     &  89.5    \\ \bottomrule
	\end{tabular}
\end{table}	
\subsubsection{Comparison with Canonical Eigenvector Sign}\label{sec:exp:aug}
As discussed in Section~\ref{sec:discuss}, the signs may bring ambiguity to \modelns. We propose to enumerate all $2^d$ eigenvectors with different signs as a data augmentation technique to solve this issue. An alternative is to use a canonical approach to determine a unique, e.g., by letting the sum of all values be positive~\cite{eigensign}. We compare these two approaches empirically for the point cloud classification task and report the results in Table~\ref{tab:cano}, where fixing the sign is denoted as \modelns-F. The results indicate that enumerating the $2^d$ possible signs can consistently improve the performance, with the cost of increasing computations. A plausible explanation is that there are potential direction issues for the canonical approach. For example, consider point clouds all representing airplanes. Though the canonical approach ensures that each airplane has a unique direction, there is no guarantee that different airplanes are aligned to the same direction, e.g., some airplanes may point towards the left, others may point towards the right, upside down, etc. The direction issue may harm the model learning. Besides, the canonical approaches can have failure cases (e.g., the sum of all values is zero), though we do not observe such failure cases in our experiments.

\begin{table*}[t]
	\caption{The results of object classification on synthetic dataset.}
	\label{tab:synthetic}
	\centering
	\begin{tabular}{c | cccccc} \toprule
		Noise & 0 & 0.25 & 0.50 & 0.75 & 1.00 \\ \midrule
		\modelns(PointNet) & 93.33$\pm$2.14 & 94.56$\pm$1.08 & 94.33$\pm$1.13 & 91.89$\pm$1.47 & 89.67$\pm$1.43\\ \bottomrule
	\end{tabular}
\end{table*}

\subsubsection{Synthetic Dataset with Eigenvalue Multiplicity}
As discussed in Section~\ref{sec:unique}, eigenvalue multiplicity may bring challenges for \modelns. To gain more empirical insights, we conduct experiments on synthetic datasets by generating and classifying point cloud objects with symmetry, which naturally results in eigenvalue multiplicity. Specifically, we randomly generate three types of objects: cylinder, regular quadrangular prism, and regular hexagonal prism. For each category, we generate 200 objects, where 70\% is used for training and the rest for testing, and each object is represented by 512 3D points. We also randomly add Gaussian noises into the input coordinate matrix. Specifically, we adopt the poisoning attack setting, i.e., the random noises are added into both the training and testing stage. We report the average results in Table~\ref{tab:synthetic} with 5 random seeds. The results show that our proposed method works reasonably well and stable on the synthetic dataset, even with random noises, indicating that eigenvalue multiplicity does not greatly affect our model empirically.

\section{Conclusion}
In this paper, we first revisit transformation invariance of geometric data using neural networks and find that transformation invariant and distance-preserving initial representations are sufficient to solve the problem. Motivated by these findings, we propose \modelns, a straightforward and general transformation invariant neural network plug-in for geometric data. 
We prove that \model can strictly guarantee transformation invariance and is general and compatible with various architectures. Experimental results on point cloud analysis and combinatorial optimization demonstrate the effectiveness and general applicability of \modelns. 

One limitation of \model is that it can only handle similarity transformation, and we plan to study other transformations (e.g., affine transformations) in the future. It would also be interesting to test \model for more applications.

\ifCLASSOPTIONcompsoc
  \section*{Acknowledgments}
\else
  \section*{Acknowledgment}
\fi

This work was supported in part by the National Key Research and Development Program of China No. 2020AAA0106300, National Natural Science Foundation of China (No. 62250008, 62222209, 62102222, 62206149), China National Postdoctoral Program for Innovative Talents No. BX20220185 and China Postdoctoral Science Foundation No. 2022M711813. All opinions, findings, conclusions and recommendations in this paper are those of the authors and do not necessarily reflect the views of the funding agencies.

\ifCLASSOPTIONcaptionsoff
  \newpage
\fi

\bibliographystyle{IEEEtran}
\bibliography{invariant}

\begin{IEEEbiography}[{\includegraphics[width=1in,height=1.25in,clip,keepaspectratio]{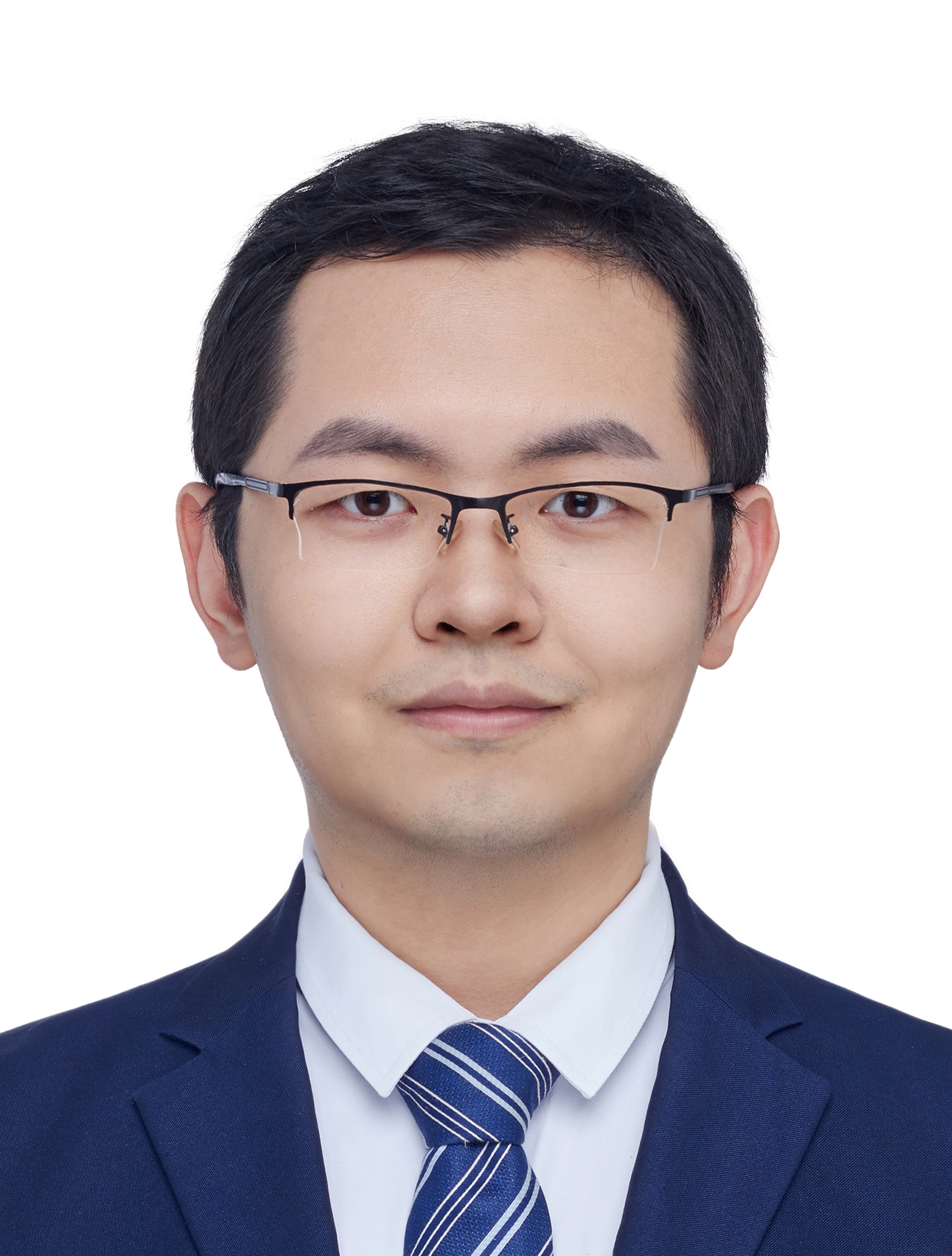}}]{Ziwei Zhang} received his Ph.D. from the Department of Computer Science and Technology, Tsinghua University, in 2021. He is currently an associate professor in the School of Computer Science and Engineering at Beihang University. His research interests focus on machine learning on graphs, including graph neural network (GNN), network embedding, and automated graph machine learning. He has published 50 papers in prestigious conferences and journals, including ACM SIGKDD, ICML, NeurIPS, AAAI, IJCAI, IEEE TPAMI, and IEEE TKDE.
\end{IEEEbiography}

\begin{IEEEbiography}[{\includegraphics[width=1in,height=1.25in,clip,keepaspectratio]{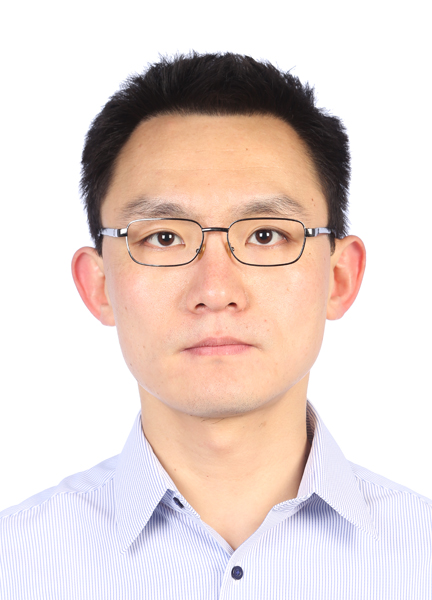}}]{Xin Wang} is currently an Assistant Professor at the Department of Computer Science and Technology, Tsinghua University. He got both of his Ph.D. and B.E degrees in Computer Science and Technology from Zhejiang University, China. He also holds a Ph.D. degree in Computing Science from Simon Fraser University, Canada. His research interests include relational media big data analysis, multimedia intelligence and recommendation in 64878853
media. He has published several high-quality research papers in top conferences including ICML, KDD, WWW, ACM Multimedia, etc. He is the recipient of 2017 China Postdoctoral innovative talents supporting program. He receives the ACM China Rising Star Award in 2020.
\end{IEEEbiography}

\begin{IEEEbiography}[{\includegraphics[width=1in,height=1.25in,clip,keepaspectratio]{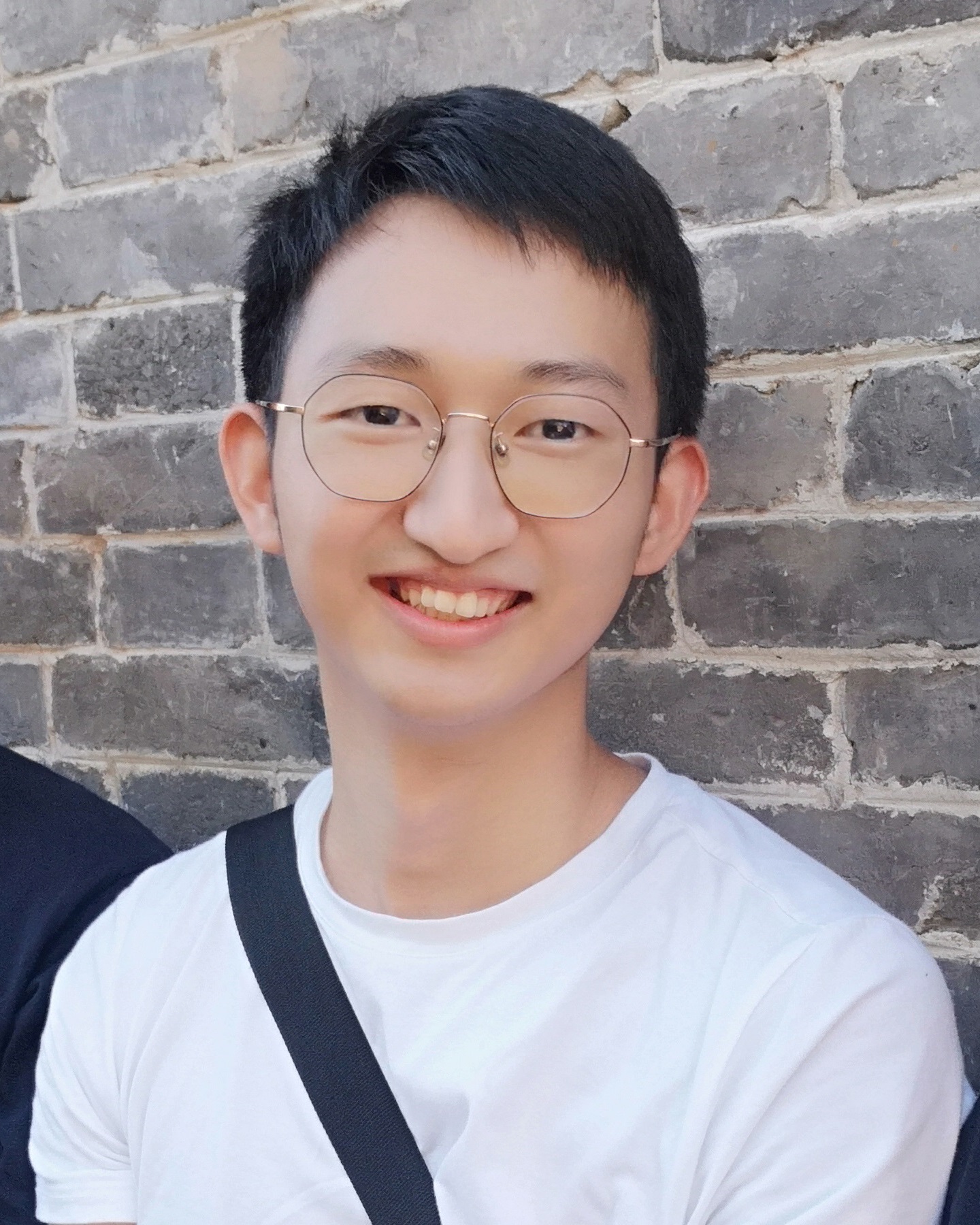}}]{Zeyang Zhang} received the B.E. from the Department of Computer Science and Technology, Tsinghua University in 2020. He is currently a Ph.D. candidate in the Department of Computer Science and Technology of Tsinghua University. His main research interests focus on graph representation learning and automated machine learning. He has published several papers in prestigious conferences, e.g., AAAI, NeurIPS, etc.
\end{IEEEbiography}

\begin{IEEEbiography}[{\includegraphics[width=1in,height=1.25in,clip,keepaspectratio]{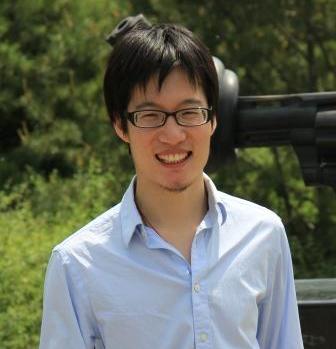}}]{Peng Cui} is an Associate Professor with tenure in Tsinghua University. He got his PhD degree from Tsinghua University in 2010. His research interests include causally-regularized machine learning, network representation learning, and social dynamics modeling. He has published more than 100 papers in prestigious conferences and journals in data mining and multimedia. His recent research won the IEEE Multimedia Best Department Paper Award, SIGKDD 2016 Best Paper Finalist, ICDM 2015 Best Student Paper Award, SIGKDD 2014 Best Paper Finalist, IEEE ICME 2014 Best Paper Award, ACM MM12 Grand Challenge Multimodal Award, and MMM13 Best Paper Award. He is PC co-chair of CIKM2019 and MMM2020, SPC or area chair of WWW, ACM Multimedia, IJCAI, AAAI, etc., and Associate Editors of IEEE TKDE, IEEE TBD, ACM TIST, and ACM TOMM etc. He received ACM China Rising Star Award in 2015, and CCF-IEEE CS Young Scientist Award in 2018. He is now a Distinguished Member of ACM and CCF, and a Senior Member of IEEE.
\end{IEEEbiography}

\begin{IEEEbiography}[{\includegraphics[width=1in,height=1.25in,clip,keepaspectratio]{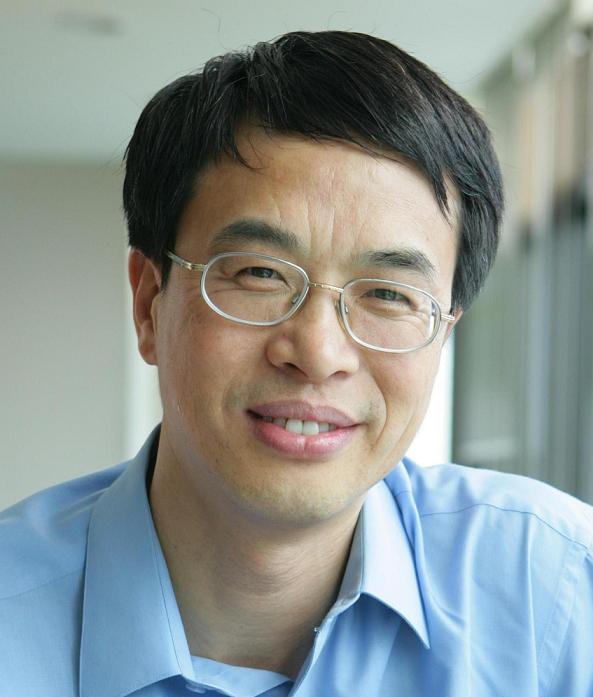}}]{Wenwu Zhu} is currently a Professor of the Computer Science Department of Tsinghua University. Prior to his current post, he was a Senior Researcher and Research Manager at Microsoft Research Asia. He was the Chief Scientist and Director at Intel Research China from 2004 to 2008. He worked at Bell Labs New Jersey as a Member of Technical Staff during 1996-1999. He received his Ph.D. degree from New York University in 1996.
	
He served as the Editor-in-Chief for the IEEE Transactions on Multimedia (T-MM) from January 1, 2017, to December 31, 2019. He has been serving as Vice EiC for IEEE Transactions on Circuits and Systems for Video Technology (TCSVT) and the chair of the steering committee for IEEE T-MM since January 1, 2020. His current research interests are in the areas of multimedia computing and networking, and big data. He has published over 350 papers in the referred journals and received nine Best Paper Awards including IEEE TCSVT in 2001 and 2019, and ACM Multimedia 2012. He is an IEEE Fellow, AAAS Fellow, SPIE Fellow and a member of the European Academy of Sciences (Academia Europaea).
\end{IEEEbiography}

\clearpage

\appendices
\section{Proofs}\label{sec:proof}
\subsection{Proof of Theorem~\ref{thm:thm1}}
\begin{proof}
	First, we show that $\tilde{\mathbf{U}}$ in Eq.~\eqref{eq:MDS1} satisfies Eq.~\eqref{eq:reqdis1}.
	
	For the Euclidean distance, we have
	\begin{equation}\label{eq:proof1}
		\begin{gathered}
			\mathcal{D}(\tilde{\mathbf{H}}^{(0)}_{i,:},\tilde{\mathbf{H}}^{(0)}_{j,:})^2 = (\tilde{\mathbf{H}}^{(0)}_{i,:} - \tilde{\mathbf{H}}^{(0)}_{j,:})(\tilde{\mathbf{H}}^{(0)}_{i,:}- \tilde{\mathbf{H}}^{(0)T}_{j,:})  \\
			= \tilde{\mathbf{H}}^{(0)}_{i,:} \tilde{\mathbf{H}}^{(0)T}_{i,:} + \tilde{\mathbf{H}}^{(0)}_{j,:} \tilde{\mathbf{H}}^{(0)T}_{j,:} - 2 \tilde{\mathbf{H}}^{(0)}_{i,:} \tilde{\mathbf{H}}^{(0)T}_{j,:}.
		\end{gathered}
	\end{equation}
	
	Using Eq.~\eqref{eq:sim2} and the eigen-decomposition, we have:
	\begin{equation}
		\tilde{\mathbf{H}}^{(0)}_{i,:} \tilde{\mathbf{H}}^{(0)T}_{j,:} = \tilde{\mathbf{S}}_{i,j}, \forall i,j.
	\end{equation}
	
	Then, we can rewrite Eq.~\eqref{eq:proof1} using Eq.~\eqref{eq:sim2} as:
	\begin{small}
		\begin{equation}
			\begin{aligned}
				& \mathcal{D}(\tilde{\mathbf{H}}^{(0)}_{i,:},\tilde{\mathbf{H}}^{(0)}_{j,:})^2 \\ &=  \tilde{\mathbf{S}}_{i,i} + \tilde{\mathbf{S}}_{j,j} - 2\tilde{\mathbf{S}}_{i,j} \\
				&=  (\mathbf{S}_{i,i} - \bar{\mathbf{S}}_{i,\cdot} - \bar{\mathbf{S}}_{\cdot,i} + \bar{\mathbf{S}}_{\cdot,\cdot}) + (\mathbf{S}_{j,j} - \bar{\mathbf{S}}_{j,\cdot} - \bar{\mathbf{S}}_{\cdot,j} + \bar{\mathbf{S}}_{\cdot,\cdot}) \\
				& \quad - 2 (\mathbf{S}_{i,j} - \bar{\mathbf{S}}_{i,\cdot} - \bar{\mathbf{S}}_{\cdot,j} + \bar{\mathbf{S}}_{\cdot,\cdot}) \\
				&=  \mathbf{S}_{i,i} + \mathbf{S}_{j,j} - 2 \mathbf{S}_{i,j} + \bar{\mathbf{S}}_{i,\cdot} - \bar{\mathbf{S}}_{j,\cdot}  + \bar{\mathbf{S}}_{\cdot,j} - \bar{\mathbf{S}}_{\cdot,i}.
			\end{aligned}
		\end{equation}
	\end{small}
	Since the Euclidean distance is symmetric, $\bar{\mathbf{S}}_{i,\cdot} = \bar{\mathbf{S}}_{\cdot,i}$ and  $\bar{\mathbf{S}}_{j,\cdot} = \bar{\mathbf{S}}_{\cdot,j}$. Besides, using Eq.~\eqref{eq:sim1}, $\mathbf{S}_{i,i} = 0$ and $\mathbf{S}_{j,j} = 0$. Thus, we have
	\begin{equation}
		\mathcal{D}(\tilde{\mathbf{H}}^{(0)}_{i,:},\tilde{\mathbf{H}}^{(0)}_{j,:})^2 = -2 \mathbf{S}_{i,j} = \mathcal{D}\left(\mathbf{F}_{i,:},\mathbf{F}_{j,:}\right)^2.
	\end{equation}
	Since both $\mathcal{D}\left(\tilde{\mathbf{H}}^{(0)}_{i,:},\tilde{\mathbf{H}}^{(0)}_{j,:}\right)$ and $\mathcal{D}\left(\mathbf{F}_{i,:},\mathbf{F}_{j,:}\right)$ are non-negative, we have $\mathcal{D}\left(\tilde{\mathbf{H}}^{(0)}_{i,:},\tilde{\mathbf{H}}^{(0)}_{j,:}\right) = \mathcal{D}\left(\mathbf{F}_{i,:},\mathbf{F}_{j,:}\right)$.
	Then, it is easy to see that
	\begin{small}
		\begin{equation}
			\mathcal{D}\left(\mathbf{H}^{(0)}_{i,:},\mathbf{H}^{(0)}_{j,:}\right) = \frac{1}{\sqrt{\mathbf{\Lambda}_{1,1}}} \mathcal{D}(\tilde{\mathbf{H}}^{(0)}_{i,:},\tilde{\mathbf{H}}^{(0)}_{j,:}) = \frac{1}{\sqrt{\mathbf{\Lambda}_{1,1}}} \mathcal{D}\left(\mathbf{F}_{i,:},\mathbf{F}_{j,:}\right)
		\end{equation}
	\end{small}
	i.e., the constant is $c^\prime = \frac{1}{\sqrt{\mathbf{\Lambda}_{1,1}}}$.
\end{proof} 
\subsection{Proof of Theorem~\ref{thm:thm3}}
\begin{proof}
	For isometric transformations in Eq.~\eqref{eq:iso}, since the distance matrix $\mathbf{D}$ is invariant by definition, the normalized similarity matrix $\tilde{\mathbf{S}}$ in Eq.~\eqref{eq:sim3} is also invariant. Thus, the point representation in Eq.~\eqref{eq:MDS2} is invariant.
	
	For the scaling transformation $\mathcal{T}\left(\mathbf{F}\right) = c\mathbf{F}$ where $c$ is an arbitrary constant, we denote all variables after the transformation with a prime in the superscripts. It is easy to see that $\mathbf{D}^\prime$ = c$\mathbf{D}$. Using Eq.~\eqref{eq:sim3}, the normalized similarity matrix satisfies:
	\begin{small}
		\begin{equation}
			\tilde{\mathbf{S}}^\prime = -\frac{1}{2} \left(\mathbf{I}_N - \frac{1}{N}\mathbf{1}_N \right) \left(\mathbf{D}^\prime \odot \mathbf{D}^\prime\right) \left(\mathbf{I}_N - \frac{1}{N}\mathbf{1}_N  \right) = c^2 \tilde{\mathbf{S}}
		\end{equation}
	\end{small}
	Using basic linear algebra knowledge, when a matrix is scaled, the eigenvalues are scales, but the eigenvectors remain unchanged, i.e., we have:
	\begin{equation}
		\mathbf{X}^\prime = \mathbf{X}, \quad \mathbf{\Lambda}^\prime = c^2 \mathbf{\Lambda}.
	\end{equation}
	Then, from Eq.~\eqref{eq:MDS2}, we have:
	\begin{equation}
		\mathbf{H}^{(0)\prime} = \mathbf{X}^\prime \sqrt{\frac{\mathbf{\Lambda}^\prime}{\mathbf{\Lambda}^\prime_{1,1}}} =  \mathbf{X} \sqrt{\frac{c^2 \mathbf{\Lambda}}{c^2 \mathbf{\Lambda}_{1,1}}} = \mathbf{H}^{(0)}.
	\end{equation}
\end{proof}		

\section{Details for Reproducibility}\label{sec:repro}
\subsection{Hyper-parameters and Detailed Settings}\label{sec:hp}
\subsubsection{Point Cloud Classification and Part Segmentation}

For the experimental settings, we exactly follow~\cite{zhang2019rotation}. In the training phase, the original data is transformed by z (in the z/z and z/SO3 setting) or SO3 (in the SO3/SO3 setting) before being fed into different models. Similarly, in the testing phase, the original data is transformed by z (in the z/SO3 setting) or SO3 (in the z/SO3 and SO3/SO3 setting) before inputting them into different models. Our proposed method adopts the same pipeline, i.e., \model receives data after the transformation, and the calculated transformation-invariant initial representations are fed into the corresponding neural networks (i.e., PointNet and DGCNN). Besides, to avoid tuning hyper-parameters, we normalize the initial representations by the average Frobenius norm of the training data so that the inputs to the neural networks are on the same scale. 

For PointNet and DGCNN, we adopt the default architectures suggested by the authors (we omit the details here, which can be found in the original paper or the implementation of the authors). For other training hyper-parameters, we also follow the suggestions in the papers. Specifically, for point cloud classification, we adopt the Adam optimizer with a learning rate of 0.001, the momentum 0.9, the decay step 200,000, and the decay rate 0.7. We train the models for 250 epochs with batch size 32 and the batch normalization with the initial decay 0.5, the decay rate 0.5, the decay step 200,000, and the decay clip 0.99. For object part segmentation, all hyper-parameters are the same except for the following: the number of training epochs is 200, the learning rate for DGCNN is 0.003, and the decay step is 675,240 for Adam and 337,620 for the batch normalization. Besides, during training for point cloud classification, PointNet also adopts a random-jittor augmentation, and DGCNN takes multiple augmentations, including random-jittor, random-scale, random-rotate-perturbation, and random-shift. We keep these augmentations unchanged as in the original model (notice that these augmentations are applied after obtaining the initial representations). For invariant baselines, we directly report their results in the paper since the same datasets split and experimental settings are adopted.

\subsubsection{Travelling Salesman Problem and Capacitated Vehicle Routing Problem} 
We closely follow~\cite{kool2018attention} for the experimental settings. Specifically, for data generation, all the coordinates of points are randomly generated in the unit square, i.e., $\left[x,y\right]$ where $x$ and $y$ are uniformly and independently drawn from $[0,1]$. For CVRP, the depot location is also randomly generated in the unit square, and the demand of point is generated as $\hat{\delta}_{i}=\frac{\delta_{i}}{D^{n}}$ where $\delta_{i}$ is sampled uniformly from $\{1, \ldots, 9\}$, and $D^{20}=30, D^{50}=40$, and $D^{100}=50$. For both efficiency and reproducibility concerns, we do not generate data on the fly. Instead, we generate 100,000 instances as the training set and 10,000 instances as the testing set. The models randomly sample data from the training set in the training phase. After training, the models are tested on the test set, and we report the average performance with standard deviation. For cases where transformations are applied, we randomly and independently apply transformations to instances in the testing set.

For the GAT architecture, the decoder structure, and training procedures, we set them exactly the same as~\cite{kool2018attention}. Specifically, the GAT has three layers with the embedding dimensionality 128 and the number of heads 8. The decoder is based on a greedy rollout trained by reinforcement learning, and the significance threshold for the paired t-test is $\alpha=0.05$. We adopt the Adam optimizer with the learning rate 1e-4, the batch size 512, the weight decay 1.0, and the number of training epochs 100.

\subsection{Source Codes and Datasets}
We adopt the following publicly available source codes and datasets.
\begin{itemize}[leftmargin=0.5cm]
	\item Point cloud experimental settings and datasets: \url{https://github.com/hkust-vgd/riconv} with MIT License
	\item PointNet: \url{https://github.com/charlesq34/pointnet}, MIT License
	\item DGCNN: \url{https://github.com/WangYueFt/dgcnn},MIT License
	\item TSP/CVRP experimental settings and baselines: \url{https://github.com/wouterkool/attention-learn-to-route}, MIT License
	\item LHK3: \url{http://akira.ruc.dk/~keld/research/LKH-3/}, license unspecified
\end{itemize}

\subsection{Software and Hardware Configurations}
All experiments are conducted on a server with the following configurations.
\begin{itemize}[leftmargin=0.5cm]
	\item Operating System: Ubuntu 18.04.1 LTS
	\item CPU: Intel(R) Xeon(R) Gold 5218 CPU @ 2.30GHz
	\item GPU: NVIDIA TESLA V100S with 32 GB of memory
	\item Software: Python 3.7.9, Cuda 10.2, PyTorch 1.7.1, TensowFlow 1.13.2, Matlab R2020b, LKH-3.0.4
\end{itemize}

\subsection{Additional Experiments on Point Cloud Analysis}
To further compare the performance of different methods, we report the results of our method and the corresponding backbones in two additional settings: none/z and none/SO3. In these settings, we do not apply data augmentations during the training phase. The results are shown in Table~\ref{tab:point:noneaug}, and we also include the results from other settings in Table~\ref{tab:classification} for reference. We make the following observations.

Under these two new settings, our method consistently and significantly outperforms the backbones, verifying that our method guarantees transformation invariance. Notably, the improvement is more significant when comparing none/z with z/z and none/SO3 with SO3/SO3, indicating that, without data augmentation, the backbones are more sensitive and vulnerable to transformations.

Additionally, when employing \modelns, data augmentation proves to be marginally beneficial for the model. Specifically, the results of \model~under z/z and SO3/SO3 are slightly higher than those under none/z and none/SO3. We attribute this enhancement to the improved generalization ability of the backbones resulting from data augmentations.More concretely, let us assume we have a table in a frontal position in the training dataset. Our model can guarantee that an identical table can be recognized from any position, including from the rear position. However, if a similar but not identical table in a rear position is present in the testing set, our method cannot strictly guarantee its recognition. On the other hand, data augmentation enables the backbone to learn what a table looks like from a rear view. Therefore, since our method is a general plug-in compatible with data augmentation, both our method and data augmentation can serve as complementary components of the model.

Additionally, we also test the inference speed of backbones and our plug-in. Specifically, for a point cloud with 2,048 points, the pre-processing step of our method takes 42.49 milliseconds, while PointNet takes 3.37 milliseconds on the ModelNet40 dataset and 4.62 milliseconds on the ShapeNet dataset. DGCNN takes 6.08 milliseconds on ModelNet40 and 8.20 milliseconds on ShapeNet. All experiments were conducted using a single Nvidia V100 GPU. Note that due to hardware changes and software upgrades, these additional results and the results in Section~\ref{sec:expscala} have slight differences. The results indicate that, although our method is slightly slower than the backbones during inference, the additional computation is not significantly dramatic.

\begin{table*}[t]
	\caption{The results of point cloud analysis on the test set. The object classification results are accuracy (\%) on the ModelNet40 dataset. The object part segmentation results are the mean per-class IoU (\%) on the ShapeNet dataset. Larger values indicate better results for both tasks. 
	}
	\label{tab:point:noneaug}
	\centering
	\begin{tabular}{c l | ccccc | ccccc } \toprule
		\multicolumn{2}{c}{Task}  & \multicolumn{5}{|c|}{Object Classification} &  \multicolumn{5}{c}{Object Part Segmentation} \\ \midrule  
		\multicolumn{2}{c|}{Setting}       		                                & z/z     &  SO3/SO3  &  z/SO3  & none/z  & none/SO3 &  z/z    &  SO3/SO3 &  z/SO3  & none/z  & none/SO3\\ \midrule
		\multirow{2}{*}{Base models}            & PointNet~\cite{qi2017pointnet}       & 87.0    &  63.6     &  13.4   & 24.8 & 10.5 & 81.0   &  71.4    &  29.0  & 38.9 & 34.7 \\ 
		& DGCNN~\cite{wang2019dynamic}         & \textbf{92.2}    &  73.3     &  22.3   & 35.7 & 18.1 & \textbf{82.0}   &  75.9    &  29.6 & 41.6 & 33.9  \\ \midrule
		\multirow{2}{*}{Our method}     & \modelns(PointNet)     			    & 86.5    &  86.5     &  86.2   & 80.2 & 79.9 & 80.9   &  80.0    &  80.0 & 72.2 & 72.0 \\
		& \modelns(DGCNN)        			    & 89.5    &  \textbf{89.5}     &  \textbf{89.5}   & \textbf{86.0} & \textbf{85.8} &  \textbf{82.0}   &  \textbf{82.1}    &  \textbf{82.0}  & \textbf{79.9} & \textbf{79.8} \\ \bottomrule
	\end{tabular}
\end{table*}

\section{Extension to Equivariance}\label{sec:equ}
Thanks to the distance preserving property of \modelns, we can extend the method into a transformation equivariant model by reversing the calculation of initial point representations and applying it at the output layer. In this section, we detail such an extension. First, we give the definition of transformation equivariance by extending Definition~\ref{def:inv}.
\begin{definition}[Transformation Equivariance]\label{def:equiv}
	For a given transformation $\mathcal{T}(\cdot):\mathbb{R}^d \rightarrow \mathbb{R}^{d}$, a neural network following Eq.~\eqref{eq:geomnn} is transformation equivariant if $\forall$ $\mathbf{F}$, $\mathbf{A}$, $\mathbf{W}$, the following equation holds:
	\begin{equation}
		\rm{NN}\left(\mathcal{T}(\mathbf{F}),\mathbf{A};\mathbf{W}\right) = \mathcal{T}\left(\rm{NN}\left(\mathbf{F},\mathbf{A};\mathbf{W}\right)\right).
	\end{equation}
\end{definition}
In a nutshell, the output point representation should be equivalently transformed as the input data. Transformation equivariance is important for many geometric deep learning tasks such as object detection and protein docking.

To extend \model into equivariance, our key idea is to find and utilize the inverse transformation of $\mathcal{P}(\cdot)$ for the points $\mathbf{F}$, which we denote as $\mathcal{P}_{\mathbf{F}}^{-1}(\cdot)$, i.e., $\mathcal{P}_{\mathbf{F}}^{-1} (\mathcal{P}(\mathbf{F})) = \mathbf{F}$. First, we introduce how to obtain $\mathcal{P}_{\mathbf{F}}^{-1}(\cdot)$. The goal of finding $\mathcal{P}_{\mathbf{F}}^{-1}(\cdot)$ is essentially finding a function to align the two sets of points, $\mathbf{F}$ and $\mathcal{P}(\mathbf{F})$, which has been extensively studied in the literature. Therefore, we apply the off-the-shelf Kabsch-Umeyama algorithm~\cite{umeyama1991least}, which can find the optimal similarity transformation that minimizes the root-mean-square deviation of the point pairs. Notice that since we have shown $\mathcal{P}(\mathbf{F})$ is fully distance-preserving, $\mathcal{P}_{\mathbf{F}}^{-1}(\cdot)$ is guaranteed to exist, i.e., the results of Kabsch-Umeyama algorithm will perfectly reverse $\mathcal{P}(\cdot)$.

After obtaining $\mathcal{P}_{\mathbf{F}}^{-1}(\cdot)$, the extension of \model into equivariance is straight-forward. First, we calculate transformation invariant point representations and conduct the forward calculation of neural networks (e.g., message-passings) as in \modelns. Next, we apply $\mathcal{P}_{\mathbf{F}}^{-1}(\cdot)$ to the outputs of the neural networks (denoted as $\hat{\mathbf{H}}^{(L)}$) to obtain final transformation equivariant point representations (denoted as $\mathbf{H}^{(L)}$). We summarize the procedures in Algorithm~\ref{alg:MDS2}. We show in the following theorem that the obtained point representations are transformation equivariant.

\begin{theorem}
    When $\mathcal{D}(\cdot,\cdot)$ is the Euclidean distance, the final representation obtained in Algorithm~\ref{alg:MDS2} satisfies Definition~\ref{def:equiv}, i.e., $\mathbf{H}^{(L)}$ is equivariant to any $\mathcal{T}(\cdot)$ in Definition~\ref{def:trans}, if $\text{rank}(\mathbf{F}) \geq d$ and the neural network is deterministic.
\end{theorem}
\begin{proof}
    For any coordinate matrix $\mathbf{F}$ and any transformed coordinate $\mathbf{F}^\prime = \mathcal{T}(\mathbf{F})$, we denote its representations in the neural network as $\mathbf{H}$ and $\mathbf{H}^\prime$. We will show that $\mathbf{H}^{(L)\prime} = \mathcal{T}(\mathbf{H}^{(L)})$.

    From Theorem~\ref{thm:thm3}, it is easy to see that $\mathcal{P}(\mathbf{F}) = \mathcal{P}(\mathbf{F}^\prime)$. Then, we have 
    \begin{equation} \mathcal{P}^{-1}_{\mathbf{F}^\prime}\left(\mathcal{P}(\mathbf{F}^\prime)\right) = \mathbf{F}^\prime = \mathcal{T}(\mathbf{F})  = \mathcal{T}(\mathcal{P}^{-1}_{\mathbf{F}}\left(\mathcal{P}(\mathbf{F}^\prime)\right)).
    \end{equation}
    Since both $\mathcal{P}(\cdot)$ and $\mathcal{T}(\cdot)$ are similarity transformations and $\text{rank}(\mathbf{F}) \geq d$ by assumptions, we have
    \begin{equation}
        \mathcal{P}^{-1}_{\mathbf{F}^\prime} = \mathcal{T} \circ \mathcal{P}^{-1}_{\mathbf{F}}.
    \end{equation}
    Besides, also from Theorem~\ref{thm:thm3}, we have $\mathbf{H}^{(0)} = \mathbf{H}^{(0)\prime}$. Since the neural network is deterministic, we have $\hat{\mathbf{H}}^{(L)} = \hat{\mathbf{H}}^{(L)\prime}$. Then, we have 
    \begin{equation}
    \mathbf{H}^{(L)\prime} = \mathcal{P}^{-1}_{\mathbf{F}^\prime}(\hat{\mathbf{H}}^{(L)\prime}) = \mathcal{T} \circ \mathcal{P}^{-1}_{\mathbf{F}} (\hat{\mathbf{H}}^{(L)}) = \mathcal{T}(\mathbf{H}^{(L)}),
    \end{equation}
    which concludes the proof. 
\end{proof}

\begin{algorithm}[t]
	\caption{A Transformation-Equivariant Neural Network Plug-in}
	\label{alg:MDS2}
	\begin{algorithmic}[1]
		\REQUIRE The coordinate matrix $\mathbf{F}$, the distance metric $\mathcal{D}\left( \cdot,\cdot\right)$, the adjacency matrix $\mathbf{A}$ \\
		\STATE Calculate $\mathbf{H}^{(0)}$ using Algorithm~\ref{alg:MDS}
        \STATE Calculate $\mathcal{P}^{-1}_{\mathbf{F}}(\cdot)$ using Kabsch–Umeyama algorithm
		\STATE Input $\mathbf{H}^{(0)}$ into neural networks to obtain $\hat{\mathbf{H}}^{(L)}$
        \STATE Calculate $\mathbf{H}^{(L)} = \mathcal{P}^{-1}_{\mathbf{F}}\left(\hat{\mathbf{H}}^{(L)}\right)$ 
	\end{algorithmic}
\end{algorithm}

\end{sloppy}
\end{document}